\documentclass{article}

\usepackage[nonatbib, preprint]{neurips_2022}

\usepackage[utf8]{inputenc} 
\usepackage[T1]{fontenc}    
\usepackage{hyperref}       
\usepackage{url}            
\usepackage{booktabs}       
\usepackage{amsfonts}       
\usepackage{nicefrac}       
\usepackage{microtype}      
\usepackage{xcolor}         

\usepackage[square, numbers]{natbib}
\usepackage[pdftex]{graphicx}
\usepackage{wrapfig}
\usepackage[skip=2pt]{caption}
\usepackage{amsmath}

\title{Accelerated Quality-Diversity through Massive Parallelism}

\author{%
    Bryan Lim  \\
    Imperial College London \\
    London, United Kingdom \\
    \texttt{bryan.lim16@imperial.ac.uk} \\
    \And
    Maxime Allard \\
    Imperial College London \\
    London, United Kingdom \\
    \texttt{m.allard20@imperial.ac.uk} \\
    \And
    Luca Grillotti \\
    Imperial College London \\
    London, United Kingdom \\
    \texttt{luca.grillotti16@imperial.ac.uk} \\
    \And
    Antoine Cully \\
    Imperial College London \\
    London, United Kingdom \\
    \texttt{a.cully@imperial.ac.uk} \\
}

\usepackage{algorithm}[H] 
\usepackage[noend]{algpseudocode}
\newbox\statebox
\newcommand{\myState}[1]{%
    \setbox\statebox=\vbox{#1}%
    \edef\thealgruleheight{\dimexpr \the\ht\statebox+1pt\relax}%
    \edef\thealgruledepth{\dimexpr \the\dp\statebox+1pt\relax}%
    \ifdim\thealgruleheight<.75\baselineskip
        \def\thealgruleheight{\dimexpr .75\baselineskip+1pt\relax}%
    \fi
    \ifdim\thealgruledepth<.25\baselineskip
        \def\thealgruledepth{\dimexpr .25\baselineskip+1pt\relax}%
    \fi
    \State #1%
    \def\thealgruleheight{\dimexpr .75\baselineskip+1pt\relax}%
    \def\thealgruledepth{\dimexpr .25\baselineskip+1pt\relax}%
}

\usepackage{stmaryrd}
\usepackage{etoolbox}
\usepackage{multirow}

\DeclareMathOperator*{\maximize}{maximize}
\begin{document}

\maketitle

\begin{abstract}
\looseness=-1 Quality-Diversity (QD) optimization algorithms are a well-known approach to generate large collections of diverse and high-quality solutions.
However, derived from evolutionary computation, QD algorithms are population-based methods which are known to be data-inefficient and require large amounts of computational resources.
This makes QD algorithms slow when used in applications where solution evaluations are computationally costly.
A common approach to speed up QD algorithms is to evaluate solutions in parallel, for instance by using physical simulators in robotics. Yet, this approach is limited to several dozen of parallel evaluations as most physics simulators can only be parallelized more with a greater number of CPUs. 
With recent advances in simulators that run on accelerators, thousands of evaluations can now be performed in parallel on single GPU/TPU.
In this paper, we present QDax, an accelerated implementation of MAP-Elites which leverages massive parallelism on accelerators to make QD algorithms more accessible.
We show that QD algorithms are ideal candidates to take advantage of progress in hardware acceleration.
We demonstrate that QD algorithms can scale with massive parallelism to be run at interactive timescales without any significant effect on the performance. 
Results across standard optimization functions and four neuroevolution benchmark environments show that experiment runtimes are reduced by two factors of magnitudes, turning days of computation into minutes.
More surprising, we observe that reducing the number of generations by two orders of magnitude, and thus having significantly shorter lineage does not impact the performance of QD algorithms.
These results show that QD can now benefit from hardware acceleration, which contributed significantly to the bloom of deep learning.
\end{abstract}

\newcommand{\R}[0]{\mathbb{R}} 

\renewcommand{\vec}[1]{{\boldsymbol{{#1}}}} 
\newcommand{\mat}[1]{{\boldsymbol{{#1}}}} 

\newcommand{\archive}{\mathcal{A}}
\newcommand{\genotype}{x}
\newcommand{\params}{\vec{\theta}}
\newcommand{\variaparams}{\widetilde{\params}}
\newcommand{\policy}{\pi_\params}

\newcommand{\fitness}{f}
\newcommand{\bd}{bd}

\newcommand{\population}{\mathcal{B}}
\newcommand{\variaPop}{\widetilde{\population}}
\newcommand{\batchsize}{N_{\mathcal{B}}}

\newcommand{\sferes}{Sferes$_{\text{v2}}$}

\newcommand{\descSet}{\mathcal{D}}

\newcommand{\cell}{\textrm{cell}}
\newcommand{\indexesCells}{\mathcal{I}}
\newcommand{\indexesParams}{\mathcal{J}}

\newcommand{\expect}[1]{ \mbox{E} \left[ {#1} \right]}

  \newcommand{\interval}{\llbracket 1, \batchsize \rrbracket}

\section{Introduction}
\vspace{-2mm}
Quality-Diversity (QD) algorithms ~\citep{pugh2016quality, cully2017quality, chatzilygeroudis2021quality} have recently shown to be an increasingly useful tool across a wide variety of fields such as robotics \citep{cully2015robots, chatzilygeroudis2018reset}, reinforcement learning (RL)~\citep{ecoffet2021first}, engineering design optimization~\citep{gaier2018data}, latent space exploration for image generation~\citep{fontaine2021differentiable} and video game design~\citep{gravina2019procedural, fontaine2020illuminating, earle2021illuminating}.
Instead of optimizing for a single solution like in conventional optimization, QD optimization searches for a population of high-performing and diverse solutions. 

Adopting the QD optimization framework has many benefits. 
The diversity of solutions found in QD enables rapid adaptation and robustness to unknown environments~\citep{cully2015robots, chatzilygeroudis2018reset, kaushik2020adaptive}.
Additionally, QD algorithms are also powerful exploration algorithms. 
They have been shown to be effective in solving sparse-reward hard-exploration tasks and achieved state-of-the-art results on previously unsolved RL benchmarks~\citep{ecoffet2021first}. 
This is a result of the diversity of solutions present acting as stepping stones~\citep{clune2019ai} during the optimization process. 
QD algorithms are also useful in the design of more open-ended algorithms~\citep{stanley2017open, stanley2019open, clune2019ai} which endlessly generate their own novel learning opporunities. 
They have been used in pioneering work for open-ended RL with environment generation~\citep{wang2019poet, wang2020enhanced}.
Lastly, QD can also be used as effective data generators for RL tasks.
The motivation for this arises from the availability of large amounts of data which resulted in the success of modern machine learning. The early breakthroughs in supervised learning in computer vision have come from the availability of large diverse labelled datasets~\citep{deng2009imagenet, barbu2019objectnet}. The more recent successes in unsupervised learning and pre-training of large models have similarly come from methods that can leverage even larger and more diverse datasets that are unlabelled and can be more easily obtained by scraping the web~\citep{devlin2018bert, brown2020language}. 
As Gu et al.~\citep{gu2021braxlines} highlighted, more efficient data generation strategies and algorithms are needed to obtain similar successes in RL.

Addressing the computational scalability of QD algorithms (focus of this work) offers a hopeful path towards open-ended algorithms that endlessly generates its own challenges and solutions to these challenges.
Likewise, they can also play a significant role in this more data-centric view of RL by generating diverse and high-quality datasets of policies and trajectories both with supervision and in an unsupervised setting~\citep{cully2019autonomous, paolo2020unsupervised}.

\begin{wrapfigure}{R}{0.5\textwidth}
\vspace{-\baselineskip}
  \centering
  \includegraphics[width=\linewidth]{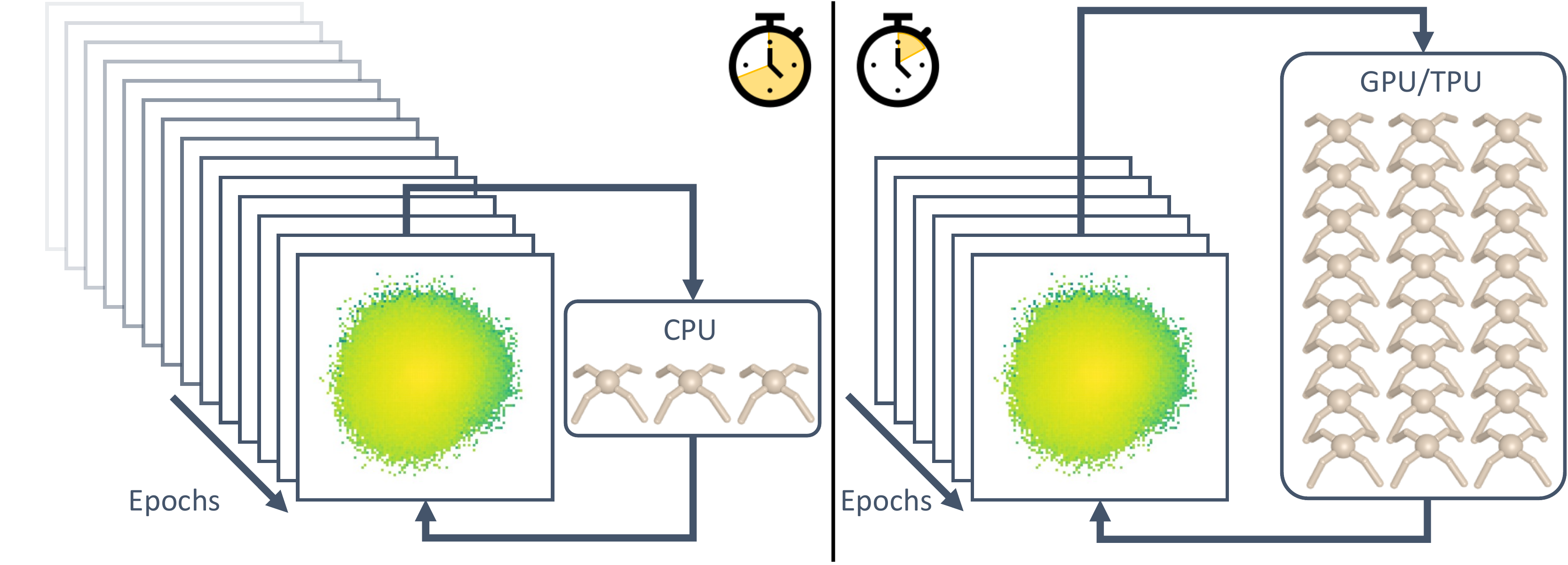}
  \caption{QDax uses massive parallelism on hardware accelerators like GPUs/TPUs to speed up runtime of QD algorithms by orders of magnitude}
  \label{fig:feature}
\end{wrapfigure}

The main bottleneck faced by QD algorithms are the large number of evaluations required that is on the order of millions.
When using QD in the field of Reinforcement Learning (RL) for robotics, this issue is mitigated by performing these evaluations in physical simulators such as Bullet~\citep{coumans2020}, DART~\citep{Lee2018}, and MuJoCo~\citep{todorov2012mujoco}.
However, these simulators have mainly been developed to run on CPUs.
Methods like MPI can be used to parallelise over multiple machines, but this requires a more sophisticated infrastructure (i.e., multiple machines) and adds some network communication overhead which can add significant runtime to the algorithm.
Additionally, the number of simulations that can be performed in parallel can only scale with the number of CPU cores available. 
Hence, both the lack of scalability coupled with the large number of evaluations required generally make the evaluation process of evolutionary based algorithms like QD, for robotics take days on modern 32-core CPUs.
Our work builds on the advances and availability of hardware accelerators, high-performance programming frameworks~\citep{jax2018github} and simulators~\citep{brax2021github} that support these devices to scale QD algorithms.

Historically, significant breakthroughs in algorithms have come from major advances in computing hardware. Most notably, the use of Graphic Processing Units (GPUs) to perform large vector and matrix computations enabled significant order-of-magnitude speedups in training deep neural networks.
This brought about modern deep-learning systems which have revolutionized  computer vision~\citep{krizhevsky2012imagenet, he2016deep, redmon2016you}, natural language processing~\citep{hochreiter1997long, vaswani2017attention} and even biology \citep{AlphaFold2021}.
Even in the current era of deep learning, significant network architectural breakthroughs~\citep{vaswani2017attention} were made possible and are shown to scale with larger datasets and more computation~\citep{devlin2018bert, shoeybi2019megatron, brown2020language, smith2022using}. 

Our goal in this paper is to bring the benefits of advances in compute and hardware acceleration to QD algorithms. The key contributions of this work are: 
(1) We show that massive parallelization with QD using large batch sizes significantly speeds up run-time of QD algorithms at no loss in final performance, turning hours/days of computation into minutes. 
(2) We demonstrate that contrary to prior beliefs, the number of iterations (generations) of the QD algorithm is not critical, given a large batch size. 
We observe this across optimization tasks and a range of QD-RL tasks.
(3) We introduce QDax, an accelerated Python implementation of MAP-Elites~\citep{mouret2015illuminating} which utilizes massive parallelization of evaluations on hardware accelerators (GPUs/TPUs). 
The source code of QDax is available at \url{https://github.com/adaptive-intelligent-robotics/QDax}.
\section{Problem Statement} 
\vspace{-2mm}

\newcommand{\statesSet}{\mathcal{S}}
\newcommand{\actionsSet}{\mathcal{A}}
\newcommand{\state}{\vec s}

\textbf{Quality-Diversity Problem.}
The Quality-Diversity (QD) problem \cite{pugh2016quality, chatzilygeroudis2021quality, fontaine2021differentiable} is an optimization problem which consists of searching for a set of solutions $\archive$ that (1) are locally optimal, and (2) exhibit diverse features.
QD problems are characterized by two components: (1) an objective function to maximize $f: \Theta \rightarrow \R$, and (2) a descriptor function $d:\Theta \rightarrow \descSet \subseteq \R^n$.
That descriptor function is used to differentiate between solutions; it takes as input a solution $\params \in \Theta$, and computes a descriptive low-dimensional feature vector.

The goal of QD algorithms is to return an \textit{archive} of solutions $\archive$ satisfying the following condition: for each achievable descriptor $\vec c \in \descSet$, there exists a solution $\params_{\archive, \vec c} \in \archive$ such that $d(\params_{\archive, \vec c}) = \vec c$ and $f(\params_{\archive, \vec c})$ maximizes the objective function with the same descriptor $\{f(\params) \mid \params\in\Theta \land d(\params) = \vec c\}$.
However, the descriptor space $\descSet$ is usually continuous.
This would require storing an infinite amount of solutions in the set.
QD algorithms alleviate this problem by considering a tesselation of the descriptor space into cells~\citep{mouret2015illuminating, cully2015robots, vassiliades2017using}: $(\cell_i)_{i\in\indexesCells}$ and keeping only a single solution per cell.
QD algorithms aim at finding a set of policy parameters $(\params_j)_{j\in\indexesParams}$ maximizing the QD-Score~\cite{pugh2016quality}, defined as follows (where $f(\cdot)$ is assumed non-negative without loss of generality):
\begin{align}
\label{eq:qd_score}
    \maximize_{(\params_j)_{j\in\indexesParams}} \quad
    \textrm{QD-Score} =  \sum_{j\in\indexesParams} f(\params_j) \quad \text{such that}\:  \forall j\in\indexesParams, \: d(\params_j) \in \cell_j
\end{align}
Thus, maximizing the QD-Score is equivalent to maximizing the number of cells containing a policy from the archive, while also maximizing the objective function in each cell.

\textbf{Quality-Diversity for Neuroevolution-based RL.} 
QD algorithms can also be used on RL problems, modeled as Markov Decision Processes $(\statesSet, \actionsSet, p, r)$, where $\statesSet$ is the states set, $\actionsSet$ denotes the set of actions, $p$ is a probability transition function, and $r$ is a state-dependent reward function.
The return $R$ is defined as the sum of rewards: $R=\sum_{t} r_t$.
In a standard RL setting, the goal is to find a policy $\pi_\params$ maximizing the expected return.

The QD for Reinforcement Learning (QD-RL) problem~\citep{nilsson2021policy, tjanaka2022approximating} is defined as a QD problem in which the goal is to find a set of diverse policy parameters $\archive = (\params_j)_{j\in\indexesParams}$ leading to diverse high-performing behaviours. 
%
In the QD-RL context, the objective function matches the expected return $f(\params_j) = \expect{ R^{(\params_j)} }$; the descriptor function $d(\cdot)$ characterizes the state-trajectory of policy $\pi_{\params_j}$: $d(\params_j)= \expect{\widetilde{d}(\vec \tau^{(\params_j)})}$ (with $\vec \tau$ denoting the state-trajectory $\vec s_{1:T}$).
The QD-Score to maximize (formula~\ref{eq:qd_score}) can then be expressed as follows, where all expected returns should be non-negative:
\begin{align}
\label{eq:qd_score_rl}
    \maximize_{(\params_j)_{j\in\indexesParams}} \quad \textrm{QD-Score} =  \sum_{j\in\indexesParams} \expect{R^{(\params_j)}} \quad \text{such that}\:  \forall j\in\indexesParams, \: \expect{ \widetilde{d}(\vec \tau^{(\params_j)}) } \in \cell_j
\end{align}

\section{Background: MAP-Elites} \label{subsec:methods-ME}
\vspace{-2mm}

\setlength{\intextsep}{0pt}%
\begin{wrapfigure}[20]{r}{0.5\textwidth}
\begin{minipage}{\linewidth}
\renewcommand*\footnoterule{}
\begin{algorithm}[H]
  \small
  \caption{MAP-Elites ($\batchsize$: Batch size)}
  \label{algo:map-elites}
  
  \newcommand\algorithmicitemindent{\hspace*{\algorithmicindent}\hspace*{\algorithmicindent}}
  \newcommand{\normal}{\mathcal{N}}
  
  \algnewcommand\algorithmicforeach{\textbf{for each}}
\algdef{S}[FOR]{ForEach}[1]{\algorithmicforeach\ #1\ \algorithmicdo}

  \begin{algorithmic}[1]
  
\For{$\textrm{iteration}\in\llbracket 1, I \rrbracket$}
\If{first iteration}
    \State{$\population\leftarrow$ random solutions}
\Else
    \State{\parbox[t]{\dimexpr\linewidth-\algorithmicindent}{$\population \leftarrow$ select solutions from archive $\archive$\strut}}
\EndIf

\State{$\variaPop = (\variaparams_j)_{j\in\interval} \leftarrow \Call{variation}{\population}$}
\State{\parbox[t]{\dimexpr\linewidth-\algorithmicindent}{$\forall j\in  \interval,$ run episode of $\pi_{\variaparams_j}$, compute return $R^{(\params_j)}$ and trajectory $\vec \tau^{(\params_j)}$\strut}}
\For{$j\in\interval$}
    \State{$\textrm{cell}  \leftarrow$ get grid cell of descriptor $\widetilde{d}(\vec \tau^{(\params_j)})$}
    \State{$\params_{\textrm{cell}} \leftarrow$ get content of cell}
    \If{$\params_{\textrm{cell}}$ is None}
    \State{Add $\params_j$ to $\textrm{cell}$}
    \ElsIf{$R^{(\params_j)} > R^{(\params_{\textrm{cell}})}$}
    \State{Replace $\params_{\textrm{cell}}$ with $\params_j$ in $\textrm{cell}$}
    \Else
    \State{Discard $\params_j$}
    \EndIf
\EndFor
\EndFor
\Return{archive $\archive$}
  \end{algorithmic}
\end{algorithm}
\end{minipage}
\end{wrapfigure}

MAP-Elites \cite{mouret2015illuminating} is a well-known QD algorithm which considers a descriptor space discretized into grid cells (Fig.~\ref{fig:all_envs_results} $4^{\textrm{th}}$ column).
At the start, an archive $\archive$ is created and initialized by evaluating random solutions. 
Then, at every subsequent iteration, MAP-Elites (i) generates new candidate solutions, (ii) evaluates their return and descriptor and (iii) attempts to add them to the archive $\archive$. The iterations are done until we reach a total budget $H$ of evaluations.
During step~(i), solutions are selected uniformly from the archive $\archive$, and undergo variations to obtain a new batch of solutions $\variaPop$. 
%
In all our experiments, we use the iso-line variation operator~\citep{vassiliades2018iso} (Appendix Algo.~\ref{algo:iso-line}).
%
%
%
Then~(ii), the solutions in the sampled batch $\variaPop = (\variaparams_j)_{j\in\interval}$ are evaluated to obtain their respective returns~$( R^{(\variaparams_j)} )_{j\in\interval}$ and descriptors~$( d(\variaparams_j) )_{j\in\interval}$. 
Finally~(iii), each solution $\variaparams_j\in\archive$ is placed in its corresponding cell in the behavioural grid according to its descriptor $d(\variaparams_j)$. 
If the cell is empty, the solution is added to the archive.
If the cell is already occupied by another solution, the solution with the highest return is kept, while the other is discarded. 
A pseudo-code for MAP-Elites is presented in Algo.~\ref{algo:map-elites}.

We use MAP-Elites to study and show how QD algorithms can be scaled through parallelization. 
We leave other variants and enhancements of QD algorithms which use learned descriptors~\citep{cully2019autonomous, paolo2020unsupervised} and different optimization strategies such as policy gradients \citep{nilsson2021policy, pierrot2021diversity, tjanaka2022approximating} and evolutionary strategies~\citep{colas2020scaling, fontaine2020covariance} for future work; we expect these variants to only improve performance on tasks, and benefit from the same contributions and insights of this work.

\section{Leveraging Hardware Acceleration for Population-based Methods} \label{sec:hardware_acc_pop_methods}
\vspace{-2mm}

In population-based methods, new solutions are the result of older solutions that have undergone variations throughout an iterative process as described in Algo. \ref{algo:map-elites}.
The number of iterations $I$ of a method commonly depends on the total number of evaluations (i.e. computational budget) $H$ for the optimization algorithm and the batch size $\batchsize$. 
For a fixed computation budget, a large batch size $\batchsize$ would result in a lower number of iterations $I$ and vice versa.
At each iteration, a single solution can undergo a variation and each variation is a learning step to converge towards an optimal solution. 
It follows that the number of iterations $I$ defines the maximum number of learning steps a solution can take. In the extreme case of $\batchsize = H$, the method simply reduces to a single random variation to a random sample of parameters.

Based on this, an initial thought would be that the performance of population-based methods would be negatively impacted by a heavily reduced number of learning steps and require more iterations to find good performing solutions.
The iterations of the algorithm are a sequential operation and cannot be parallelized. 
However, the evaluation of solutions at each iteration can be massively parallelized by increasing the batch size $\batchsize$ which suits modern hardware accelerators.
We investigate the effect of large $\batchsize$ on population-based methods (more specifically QD algorithms) by ablating exponentially increasing $\batchsize$ which was relatively unexplored in the literature.

Conventional QD algorithms parallelize evaluations by utilizing multiple CPU cores, where each CPU separately runs an instance of the simulation to evaluate a solution. 
For robotics experiments, we utilize Brax~\citep{brax2021github}, a differentiable physics engine in Python which enables massively parallel rigid body simulations. By leveraging a GPU/TPU, utilizing this simulator allows us to massively parallelize the evaluations in the QD loop which is the major bottleneck of QD algorithms. 
To provide a sense of scale, QD algorithms normally run on the order of several dozens of evaluations in parallel with $\batchsize \sim 10^2$ due to limited CPUs while Brax can simulate over 10,000 solutions in parallel allowing QDax to have $\batchsize \sim 10^5$.
Brax is built on top of the JAX~\citep{jax2018github} programming framework, which provides an API to run accelerated code across any number of hardware acceleration devices such as CPU, GPU or TPU. 

Beyond massive parallelization, acceleration of our implementation is also enabled with code compatible with just-in-time (JIT) compilation and fully on-device computation. 
Another key benefit of JAX is the JIT compilation which allows JAX to make full use of its Accelerated Linear Algebra (XLA).We provide implementation details with static archives that makes QDax compatible with JIT in the Appendix.
Lastly, another bottleneck which slowed the algorithm down was the data transfer and marshalling across devices. 
To address this issue, we carefully consider data structures and place all of them on-device. 
QDax places the JIT-compiled QD algorithm components on the same device.
This enables the entire QD algorithm to be run without interaction with the CPU.
\\
\begin{figure}[h]
  \centering
  \includegraphics[width=\linewidth]{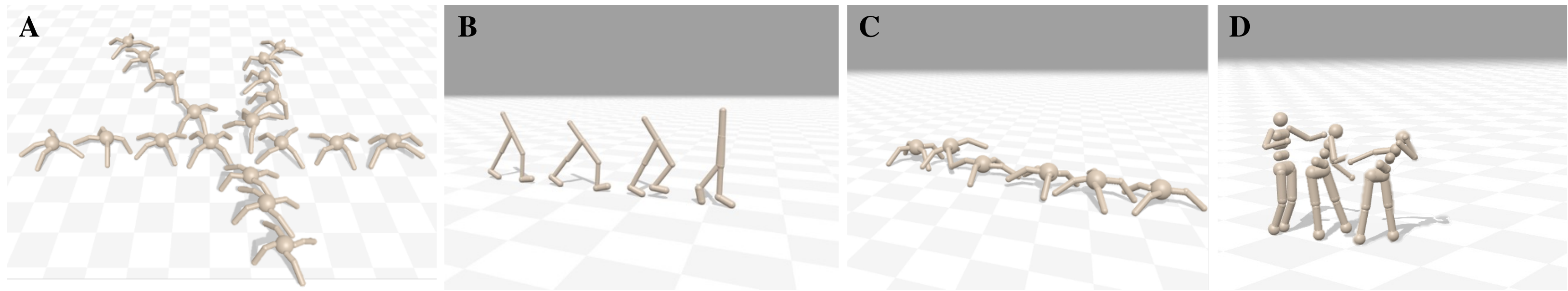}
  \caption{QD algorithms can discover diverse solutions. A: Omni-directional Ant task discovers ways to move in every direction. B,C,D: Uni-directional Walker, Ant and Humanoid task respectively discover diverse gaits for moving forward.}
  \label{fig:behaviours}
  \vspace{-2mm}
\end{figure}

\vspace{-1mm}
\section{Experiments}
\vspace{-2mm}
\label{sec:experiments}
\looseness=-1 Our experiments aim to answer the following questions: 
(1) How does massive parallelization affect performance of Quality-Diversity (MAP-Elites) algorithms? 
(2) How significant is the number of iterations/learning steps in QD algorithms? 
(3) What magnitude of speed-up does massive parallelization offer over existing implementations? 
(4) How does this differ across different hardware accelerators? 

\subsection{Domains}
\vspace{-1mm}
\looseness=-1 \textbf{Rastrigin and Sphere.}
Rastrigin and Sphere are standard optimization functions commonly used as benchmark domains in optimization~\citep{hansen:hal-00545727, hansen2021coco} and QD-literature~\citep{fontaine2020covariance, fontaine2021differentiable}. We optimize a $n=100$ dimensional parameter space bounded between 0 and 1. More details regarding the objective function and descriptors are provided in the Appendix. We select these simple domains to demonstrate that the phenomena of large batch sizes applies generally to QD algorithms and not just to tasks in certain domains.

\textbf{Continuous Control QD-RL.}
We perform experiments on two different QD-RL benchmark tasks~\citep{cully2015robots, nilsson2021policy, tjanaka2022approximating}; \textit{omni-directional} robot locomotion and \textit{uni-directional} robot locomotion.
In the omni-directional task, the goal is to discover locomotion skills to move efficiently in every direction.
The descriptor functions is defined as the final $x$-$y$ positions of the centre of mass of the robot at the end of the episode while the objective $f$ is defined as a sum of a survival reward and torque cost.
In contrast, the goal in the uni-directional tasks is to find a collection of diverse gaits to walk forward as fast as possible. In this task, the descriptor function is defined as the average time over the entire episode that each leg is in contact with the ground. For each foot $i$, the contact with the ground $C_i$ is logged as a Boolean (1: contact, 0: no-contact) at each time step $t$. The descriptor function of this task was used to allow robots to recover quickly from mechanical damage~\citep{cully2015robots}. The objective $f$ of this task is a sum of the forward velocity, survival reward and torque cost.
Full equation details can be found in the Appendix.

We use the Walker2D, Ant and Humanoid gym locomotion environments made available on Brax~\citep{brax2021github} on these tasks.
We run the omni-directional experiments for an episode length $T$ of 100 steps and uni-directional experiments for 300 steps.
In total, we report results on a combination of four tasks and environments; Omni-directional Ant and Uni-directional Walker, Ant and Humanoid.
Fig.~\ref{fig:behaviours} illustrates examples of the types behaviors discovered from these tasks.
We use a grid discretizations of shape [100,100], [10,10,10,10], [40,40] and [40.40] respectively for the corresponding tasks.
We use fully connected neural network controllers with two hidden layers of size 64 and tanh output activation functions as policies across all environments and tasks.


\subsection{Effects of massive parallelization on QD algorithms} \label{subsec:eff_of_massive_par}
\vspace{-1mm}
To evaluate the effect of the batchsize $\batchsize$, we run the algorithm for a fixed number of evaluations. 
We use 5 million evaluations for all QD-RL environments and 20 million evaluations for \textit{rastrigin} and \textit{sphere}.
We evaluate the performance of the different batch sizes using the QD-score.
The QD-score~\citep{pugh2016quality} aims to capture both performance and diversity in a single metric. This metric is computed as the sum of objective values of all solutions in the archive (Eqn.~\ref{eq:qd_score}).
We plot this metric with respect to three separate factors: number of evaluations, number of iterations and total runtime.
Other commonly used metrics in QD literature such as the best objective value and coverage can be found in the Appendix.
We use a single A100 GPU to perform our experiments. 

\begin{figure*}[h]
  \centering
  \includegraphics[width=\linewidth]{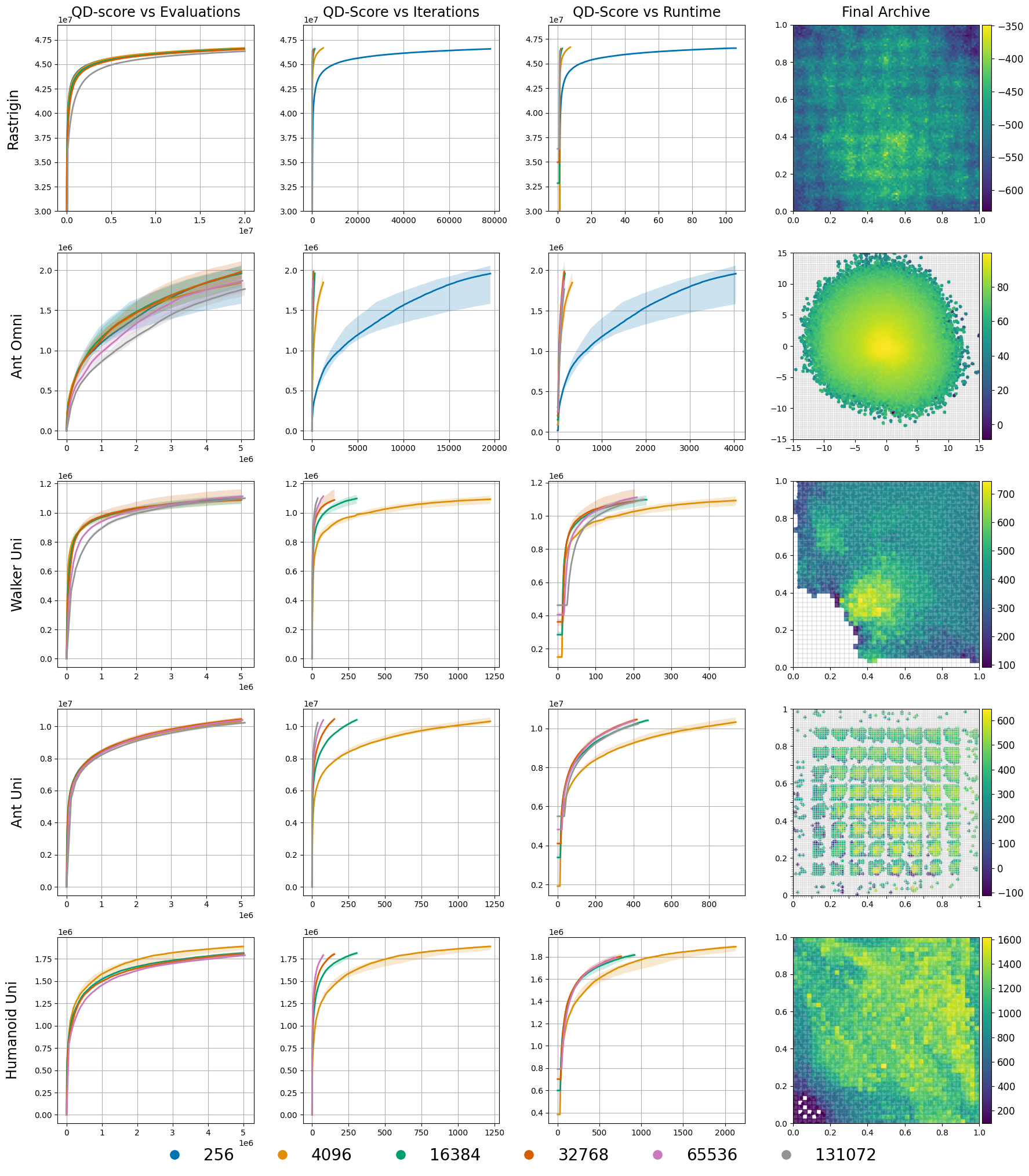}
  \caption{Performance metrics across domains. The plots show the QD-Score against the number of evaluations, iterations and run-times for each batch size. The rightmost column shows the final archives.
  For Ant Uni, the 4D descriptors are shown in 2D as done in (Cully et al.)~\cite{cully2015robots}.
  The bold lines and shaded areas represent the median and interquartile range over 10 replications respectively.
  We leave the $\batchsize=256$ out of the Uni environments to increase readability of the large batch sizes.}
  \label{fig:all_envs_results}
  \vspace{-4mm}
\end{figure*}

Fig.~\ref{fig:all_envs_results} shows the performance curves for QD-score and more importantly the differences when plot against the number of evaluations, iterations and total runtime.
A key observation in the first column is that the metrics converge to the same final score after the fixed number of evaluations regardless of the batch size used. 
The Wilcoxon Rank-Sum Test for the final QD score across all the different batches results in p-values p>0.05 after applying the Bonferroni Correction. This shows that we do not observe statistically significant differences between the different batch sizes. Therefore, larger batch sizes and massive parallelism do not negatively impact the final performance of the algorithm.
However, an important observation is that the larger batch sizes have a trend to be slower in terms of number of evaluations to converge. This can be expected as a larger number of evaluations are performed per iteration at larger batch sizes.
Conversely, in some cases (Ant-Uni), a larger batch size can even have a positive impact on the QD-score.
Given this result, the third column of Fig. \ref{fig:all_envs_results} then demonstrates the substantial speed-up in total runtime of the algorithm obtained from using larger batch sizes with massive parallelism while obtaining the same performances.
We can obtain similar results but in the order of minutes instead of hours.
An expected observation we see when comparing the plots in the second and third column of Fig. \ref{fig:all_envs_results} is how the total runtime is proportional to the number of iterations. As we are able to increase the evaluation throughput at each iteration through the parallelization, it takes a similar amounts of time to evaluate both smaller and larger batch sizes.
The speed-up in total runtime of the algorithm by increasing batch size eventually disappears as we reach the limitations of the hardware. This corresponds to the results presented in the the next section \ref{sec:4_1} (Fig.~\ref{fig:evalpers}) where we see the number eval/s plateauing. 

\looseness=-1 The most surprising result is that the number of iterations and thus learning steps of the algorithm do not significantly affect the performance of the algorithm when increasing batch sizes $\batchsize$ are used.
In the case of the \textit{QD-RL} domains, we observe that using $\batchsize=131{,}072$ which runs for only $I=39$ provides the same performance as when run with $\batchsize=256$ and $I=19{,}532$. This is true for all the domains presented.
Given a fixed number of evaluations, a larger batch size would imply a lower number of iterations. This can be observed in the second column of Fig. \ref{fig:all_envs_results}.
Therefore, although our results show that a larger batch size with less iterations has no negative impact on QD algorithms and can significantly speed-up the runtime of QD algorithms, the algorithm still needs to be allowed to run for a sufficient number of iterations to reach the similar final performance scores.
The iterations/learning steps remain an important part of QD algorithms as new solutions that have been recently discovered and added to the archive $\archive$ from a previous iteration can be selected as stepping stones to form future solutions.
Results from our experiments show that iterations are not important as long as the number of evaluations remains identical.
This is a critical observation as while iterations cannot be performed in parallel, evaluations can be, which enables the use of massive parallelization.

\subsection{Evaluation throughput and runtime speed of QD algorithms} \label{sec:4_1}
\vspace{-1mm}
\looseness=-1 We also evaluate the effect that increasing batch sizes have on the evaluation throughput of QD algorithms.
We start with a batch size $\batchsize$ of 64 and double from this value until we reach a plateau and observe a drop in performance in throughput. In practice, a maximum batch size of 131,072 is used.

The number of evaluations per second (eval/s) is used to quantify this throughput. The eval/s metric is computed by running the algorithm for a fixed number of generations (also referred to as iterations) $N$ (100 in our experiments). We divide the corresponding batch size $\batchsize$ representing the number of evaluations performed in this iteration and divide this value by the time it takes to perform one iteration $t_n$.
We use the final average value of this metric across the entire run: $eval/s = \frac{1}{N} \sum_{n=1}^{N} \frac{\batchsize}{t_n}$. 
While the evaluations per second can be an indication of the improvement in throughput from this implementation, we ultimately care about running the entire algorithm faster. To do this, we evaluate the ability to speed up the total runtime of QD algorithms. In this case, we run the algorithm to a fixed number of evaluations (1 million), as usually done in QD literature.
Running for a fixed number of iterations would be an unfair comparison as the experiments with smaller batch sizes would have much less evaluations performed in total. 

\begin{wrapfigure}{R}{0.5\textwidth}
  \centering
  \includegraphics[width=\linewidth]{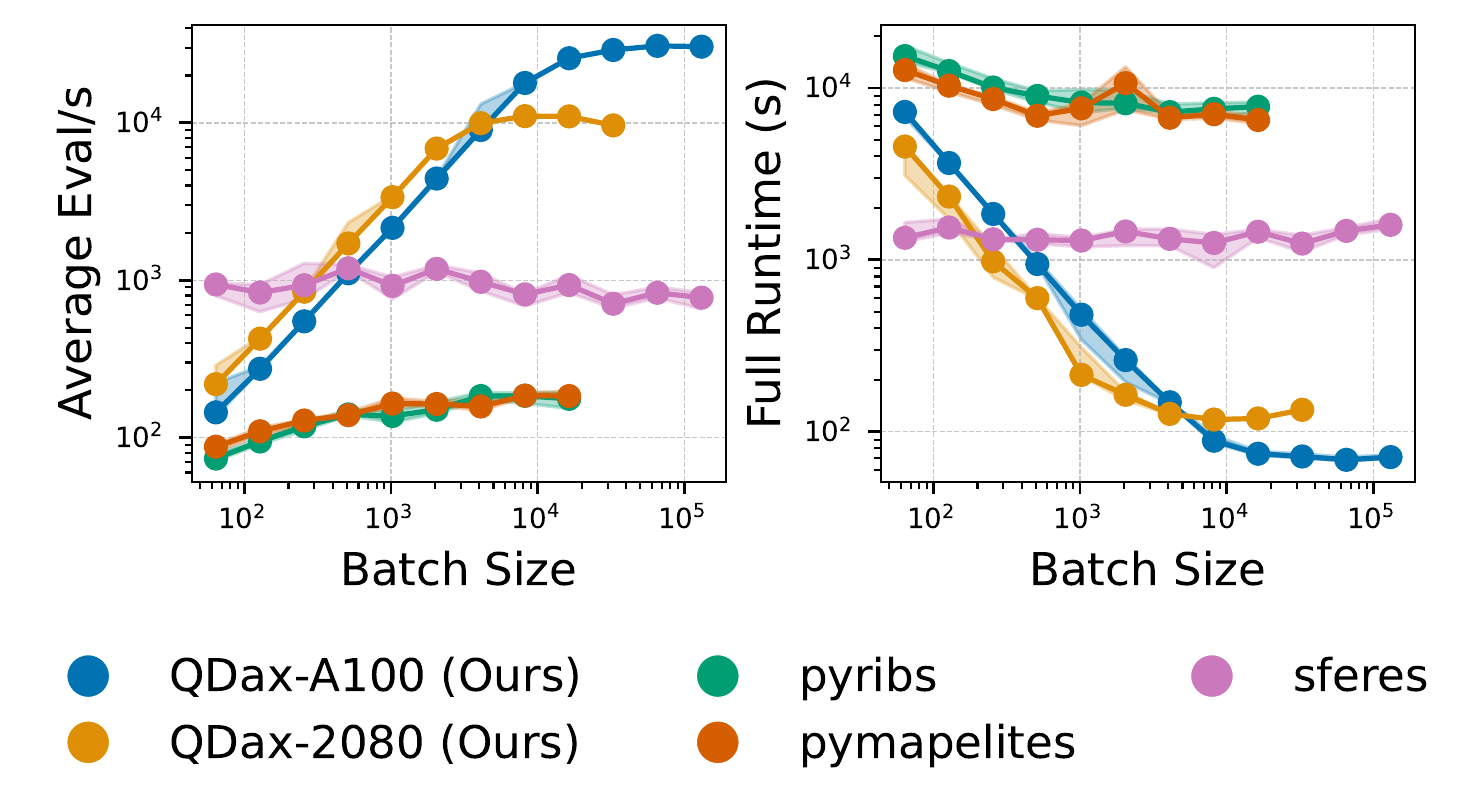}
  \caption{Average number of eval/s and full runtime of algorithm across batch sizes and implementations. Note the log scales on both axes to make distinction between batch sizes clearer.
  }
  \label{fig:evalpers}
\end{wrapfigure}
For this experiment, we consider the Ant Omni-directional task.
We compare against common implementations of MAP-Elites and open-source simulators which utilize parallelism across CPU threads (see Table \ref{tab:freq}).
All baseline algorithms used simulations with a fixed number of timesteps (100 in our experiments). 
We compare against both Python and C++ implementations as baselines. 
Pymapelites~\citep{mouret2015illuminating} is a simple reference implementation from the authors of MAP-Elites that was made to be easily transformed for individual purposes. 
Pyribs~\citep{pyribs} is a more recent QD optimization library maintained by the authors of CMA-ME~\citep{fontaine2020covariance}.
In both Python implementations, evaluations are parallelised on each core using the multiprocessing Python package.
Lastly, \sferes{}~\citep{Mouret2010} is an optimized, multi-core and lightweight C++ framework for evolutionary computation, which includes QD implementations. It relies on template-based programming to achieve optimized execution speeds. The multi-core distribution for parallel evaluations is handled by Intel Threading Building Blocks (TBB) library.
For simulators, we use PyBullet~\citep{coumans2020} in our Python baselines and Dynamic Animation and Robotics Toolkit (DART) ~\citep{Lee2018} for the C++ baseline.
We recognize that this comparison is not perfect and there could exist more optimized combinations of implementations and simulators but believe the baselines selected gives a good representation of what is commonly used in most works~\cite{cully2013behavioral, nilsson2021policy, tjanaka2022approximating}.

We also test our implementation on two different GPU devices, a more accessible RTX2080 local device and a higher-performance A100 on Google Cloud. We only consider a single GPU device at each time. QDax was also tested and can be used to perform experiments across distributed GPU and TPU devices but we omit these results for simplicity. 

\begin{table*}
\centering
\small
  \begin{tabular}{lll | rr | rr}
    \toprule
    Implementation&Simulator&Resources&
    Eval/s&Batch size&Runtime (s)&Batch size\\
    \midrule
        \addlinespace
    QDax (Ours) &Brax~\citep{brax2021github}& GPU A100&
    30,846 & 65,536 & \textbf{69}& 65,536\\
    QDax (Ours) &Brax~\citep{brax2021github}& GPU 2080&
    11,031 & 8,192 & 117 & 8,192\\
    \addlinespace
    pyribs~\citep{pyribs} & PyBullet~\citep{coumans2020} &32 CPU-cores& 
    184& 8,192& 7,234& 4,096\\
    pymapelites~\citep{mouret2015illuminating}& PyBullet~\citep{coumans2020} & 32 CPU-cores& 185& 8,192& 6,509& 16,384\\
    \sferes~\citep{Mouret2010} & DART~\citep{Lee2018} &32 CPU-cores& 1,190& 512& 1,243& 32,768\\
  \bottomrule
\end{tabular}
  \caption{Maximum throughput of evaluations per second and fastest run time obtained and their corresponding batch sizes across implementations. The medians over the 10 replications are reported.}
  \label{tab:freq}
  \vspace{-2mm}
\end{table*}

Fig.~\ref{fig:evalpers} (Left) clearly shows that QDax has the ability to scale to much larger batch sizes which results in a higher throughput of evaluations. 
It is important to note the log scale on both axes to appreciate the magnitude of the differences.
For QDax implementations (blue and orange), the number of evaluations per second scales as the batch size used increases. This value eventually plateaus once we reach the limit of the device. 
On the other hand, all the baseline implementations scales to a significantly lower extent. These results can be expected as evaluations using simulators which run on CPUs are limited as each CPU core runs a separate instance of the simulation. Therefore, given only a fixed number of CPU cores, these baselines would not scale as the batch size is increased. Scaling is only possible by increasing the number of CPUs used in parallel which is only possible with large distributed system with thousands of CPUs in the network. 
QDax can reach up to a maximum of ~30,000 evaluations per second on an A100 GPU compared to maximum of ~1,200 (C++) or ~200 (Python) evaluations per second in the baselines (see Table \ref{tab:freq}).
This is a 30 to 100 times increase in throughput, turning computation on the order of days to minutes.
The negligible differences between the pyribs (green) and pymapelites (red) results show that the major bottleneck is indeed the evaluations and simulator used, as both of these baselines use the PyBullet simulator.
The performance of the \sferes{} (purple) implementation can be attributed to its highly optimized C++ code. However, the same lack of scalability is also observed when the batchsize is increased.
When looking at run time of the algorithm for a fixed number of evaluations on Fig.~\ref{fig:evalpers} (Right), we can see the effect of the larger throughput of evaluations at each iteration reflected in the decreasing run-time when larger batch sizes are used. We can run a QD algorithm with 1 million evaluations in just slightly over a minute (See Table \ref{tab:freq}) when using a batch size of 65,536 compared to over 100 minutes taken by Python baselines.

Our results also show that this scaling through massive parallelism is only limited by the hardware available. The experiments on both the RTX2080 (orange) and A100 (blue) show similar trends and increases in both evaluations per second and total runtime. The 2080 plateaus at a batch size of 8,192 capable of ~11,000 eval/s while the higher-end A100 plateaus later at a batch size of 65,536 completing ~30,000 eval/s.

\vspace{-1mm}
\section{Related Work}
\vspace{-2mm}
\textbf{Quality-Diversity}. 
QD algorithms were derived from interests in divergent search methods~\citep{lehman2011abandoning} and behavioural diversity~\citep{mouret2009overcoming, mouret2012encouraging} in evolutionary algorithms and hybridizing such methods with the notion of fitness and reward~\citep{lehman2011evolving}. 
While QD algorithms are promising solutions to robotics and RL, they remain computationally expensive and take a long time to converge due to high sample-complexity. 
The move towards more complex environments with high-dimensional state and action spaces, coupled with the millions of evaluations required for the algorithm to converge, makes these algorithms even more inaccessible to regular hardware devices.
Progress has been made towards lowering the sample complexity of QD algorithms and can generally be categorized into two separate approaches. 
The first approach is to leverage the efficiency of other optimization methods such as evolution strategies~\citep{colas2020scaling, fontaine2020covariance, cully2020multi} and policy-gradients~\citep{nilsson2021policy, pierrot2021diversity}. 
The other line of work known as \textit{model-based quality-diversity}~\citep{gaier2018data, keller2020model, lim2021dynamics}, reduces the number of evaluations required through the use of surrogate models to provide a prediction of the descriptor and objective. 

Our work takes a separate approach orthogonal to sample-efficiency and focuses improvement on the runtime of QD algorithms instead by leveraging the batch size at each iteration.
Additionally, despite algorithmic innovations that improve sample-efficiency, most QD implementations still rely on evaluations being distributed over large compute systems. These often give impressive results~\citep{colas2020scaling, fontaine2019mapping} but such resources are mainly inaccessible to most researchers and yet still take significant amount of time to obtain.
Our work aims to make QD algorithms more accessible by running quickly on more commonly available accelerators, such as cloud available GPUs. 

\textbf{Hardware Acceleration for Machine Learning.} Machine Learning, and more specifically Deep Learning methods, have benefited from specialised hardware accelerators that can parallelize operations. In the mid-2000's researchers started using GPUs to train neural networks~\citep{steinkrau205gpu} because of their high degree of parallelism and high memory bandwidth. 
After the introduction of general purpose GPUs, the use of specialized GPU compatible code for Deep Learning methods~\citep{raina2009gpu, ciresan2012gpu} enabled deep neural networks to be trained a few orders of magnitude quicker than previously on CPUs~\citep{lecun2015DL}. Very quickly, frameworks such as Torch~\citep{collobert2011torch}, Tensorflow~\citep{abadi2016tf}, PyTorch~\citep{paszke2019pytorch} or more recently JAX~\citep{jax2018github} were developed to run numerical computations on GPUs or other specialized hardware.

These frameworks have led to tremendous progress in deep learning. In other sub-fields such as deep reinforcement learning (DRL) or robotics, the parallelization has happened on the level of neural networks.
However, RL algorithms need a lot of data and require interaction with the environment to obtain it. 
Such methods suffer from a slow data collection process as the physical simulators used to collect data were mainly developed for CPUs, which results in a lack of scalable parallelism and data transfer overhead between devices. 
More recently, new rigid body physics simulators that can leverage GPUs and run thousands of simulations in parallel have been developed. Brax~\citep{brax2021github} and IsaacGym~\citep{makoviychuk2021isaac} are examples of these new types simulators.
Gradient-based DRL methods can benefit from this massive parallelism as this would directly correspond to estimating the gradients more accurately through collection of larger amounts of data in a faster amount of time~\citep{rudin2021learning} at each optimization step.
Recent work~\citep{gu2021braxlines, rudin2021learning} show that control policies can be trained with DRL algorithms like PPO~\citep{schulman2017proximal} and DIAYN~\citep{eysenbach2018diversity} in minutes on a single GPU.
However, unlike gradient-based methods, it is unclear until now how evolutionary and population-based approaches like QD would benefit from this massive parallelism since the implications of massive batch sizes has not been studied to the best of our knowledge.
Our work studies the impact of massive parallelization on QD algorithms and its limitations.

\vspace{-2mm}
\section{Societal Impact} \label{societal-impact}
\vspace{-2mm}
Accessibility was a key consideration and motivation of this work.
Our work turns the execution of QD algorithms from what took days/weeks on large CPU clusters to only minutes on a single easily accessible and free cloud GPU/TPU.
We hope that this will prevent limitations of access of these algorithms to wider and more diverse communities of people, particularly from emerging and developing economies and beyond only well-resourced research groups and enterprises.
We believe this increase in accessibility has the potential to open up novel applications of QD in new domains and lead to algorithmic innovations. 
On the other hand, a key scientific takeaway from our work was also that QD algorithms were able to scale with more powerful and modern hardware.
We recognize that this could also open up the possibility of 'buying' results~\citep{schwartz2020green} by utilizing more compute.

We are also aware that algorithms that can scale well with compute come at the risk of carbon footprint cost.
Evident from the bloom of deep learning and more recently large language models, computational demands only continue to increase with models that scale.
Interestingly, our work shows that given the same machine, we can decrease computation time by two orders of magnitude with no loss in performance of the algorithm which can aid in reducing environmental impact.
Nonetheless, we highlight research on the environmental and carbon impact of AI and recommendations to be accountable in minimizing its effects~\citep{dobbe2019climate, dhar2020carbon}.
A common step towards reducing AI's climate impact is to increase transparency on energy consumption and carbon emissions from computational resources used.
We report the estimated emissions used in our experiments in the Appendix.
Estimations were conducted using the \href{https://mlco2.github.io/impact#compute}{Machine Learning Impact calculator}~\citep{lacoste2019quantifying}.

\vspace{-2mm}
\section{Limitations and Future Work} \label{sec:discussion}
\vspace{-2mm}
In this paper, we presented QDax, an implementation of MAP-Elites that utilizes massive parallelization on accelerators that reduce the runtime of QD algorithms to interactive timescales on the order of minutes instead of hours or days. We evaluate QDax across a range QD tasks and show that the performance of QD algorithms are maintained despite the significant speed-up that comes with the massive parallelism.
Despite the apparent importance of iterations in QD algorithms, we show that when large batch sizes are used, a heavily reduced number of iterations and hence learning steps provides similar results.

Despite reaping the benefits of hardware in order to accelerate the algorithm, there are still some limitations in the future. 
As the archive stores the parameters among other things, the memory of the device becomes an issue preventing larger networks with more parameters from being used. 
Similarly, this memory limitation also prevent larger archives with more cells from being used. 
However, the experiments in this paper do not do use anything less than what is commonly used in literature. 

Through this work, we hope to increase accessibility of QDax can help bring ideas from an emerging field of optimization to accelerate progress in machines learning. We also hope to see new algorithmic innovations that could leverage the massive parallelization to improve performance of QD algorithms.

\begin{ack}


This work was supported by the Engineering and Physical Sciences Research Council (EPSRC) grant EP/V006673/1 project REcoVER, and by Google with GCP credits. We would like to thank the members of the Adaptive and Intelligent Robotics Lab for their very valuable comments. 
\end{ack}

{
\small
\bibliographystyle{plainnat}
\bibliography{references}

\begin{thebibliography}{73}
\providecommand{\natexlab}[1]{#1}
\providecommand{\url}[1]{\texttt{#1}}
\expandafter\ifx\csname urlstyle\endcsname\relax
  \providecommand{\doi}[1]{doi: #1}\else
  \providecommand{\doi}{doi: \begingroup \urlstyle{rm}\Url}\fi

\bibitem[Abadi et~al.(2016)Abadi, Barham, Chen, Chen, Davis, Dean, Devin,
  Ghemawat, Irving, Isard, Kudlur, Levenberg, Monga, Moore, Murray, Steiner,
  Tucker, Vasudevan, Warden, Wicke, Yu, and Zheng]{abadi2016tf}
Martin Abadi, Paul Barham, Jianmin Chen, Zhifeng Chen, Andy Davis, Jeffrey
  Dean, Matthieu Devin, Sanjay Ghemawat, Geoffrey Irving, Michael Isard,
  Manjunath Kudlur, Josh Levenberg, Rajat Monga, Sherry Moore, Derek~G. Murray,
  Benoit Steiner, Paul Tucker, Vijay Vasudevan, Pete Warden, Martin Wicke, Yuan
  Yu, and Xiaoqiang Zheng.
\newblock Tensorflow: A system for large-scale machine learning.
\newblock In \emph{12th USENIX Symposium on Operating Systems Design and
  Implementation (OSDI 16)}, pages 265--283, 2016.
\newblock URL
  \url{https://www.usenix.org/system/files/conference/osdi16/osdi16-abadi.pdf}.

\bibitem[Barbu et~al.(2019)Barbu, Mayo, Alverio, Luo, Wang, Gutfreund,
  Tenenbaum, and Katz]{barbu2019objectnet}
Andrei Barbu, David Mayo, Julian Alverio, William Luo, Christopher Wang, Danny
  Gutfreund, Joshua Tenenbaum, and Boris Katz.
\newblock Objectnet: A large-scale bias-controlled dataset for pushing the
  limits of object recognition models.
\newblock 2019.

\bibitem[Bradbury et~al.(2018)Bradbury, Frostig, Hawkins, Johnson, Leary,
  Maclaurin, Necula, Paszke, Vander{P}las, Wanderman-{M}ilne, and
  Zhang]{jax2018github}
James Bradbury, Roy Frostig, Peter Hawkins, Matthew~James Johnson, Chris Leary,
  Dougal Maclaurin, George Necula, Adam Paszke, Jake Vander{P}las, Skye
  Wanderman-{M}ilne, and Qiao Zhang.
\newblock {JAX}: composable transformations of {P}ython+{N}um{P}y programs,
  2018.
\newblock URL \url{http://github.com/google/jax}.

\bibitem[Brown et~al.(2020)Brown, Mann, Ryder, Subbiah, Kaplan, Dhariwal,
  Neelakantan, Shyam, Sastry, Askell, et~al.]{brown2020language}
Tom~B Brown, Benjamin Mann, Nick Ryder, Melanie Subbiah, Jared Kaplan, Prafulla
  Dhariwal, Arvind Neelakantan, Pranav Shyam, Girish Sastry, Amanda Askell,
  et~al.
\newblock Language models are few-shot learners.
\newblock \emph{arXiv preprint arXiv:2005.14165}, 2020.

\bibitem[Chatzilygeroudis et~al.(2018)Chatzilygeroudis, Vassiliades, and
  Mouret]{chatzilygeroudis2018reset}
Konstantinos Chatzilygeroudis, Vassilis Vassiliades, and Jean-Baptiste Mouret.
\newblock Reset-free trial-and-error learning for robot damage recovery.
\newblock \emph{Robotics and Autonomous Systems}, 100:\penalty0 236--250, 2018.

\bibitem[Chatzilygeroudis et~al.(2021)Chatzilygeroudis, Cully, Vassiliades, and
  Mouret]{chatzilygeroudis2021quality}
Konstantinos Chatzilygeroudis, Antoine Cully, Vassilis Vassiliades, and
  Jean-Baptiste Mouret.
\newblock Quality-diversity optimization: a novel branch of stochastic
  optimization.
\newblock In \emph{Black Box Optimization, Machine Learning, and No-Free Lunch
  Theorems}, pages 109--135. Springer, 2021.

\bibitem[Ciresan et~al.(2012)Ciresan, Meier, and Schmidhuber]{ciresan2012gpu}
Dan~C. Ciresan, Ueli Meier, and J{\"{u}}rgen Schmidhuber.
\newblock Multi-column deep neural networks for image classification.
\newblock \emph{CoRR}, abs/1202.2745, 2012.
\newblock URL \url{http://arxiv.org/abs/1202.2745}.

\bibitem[Clune(2019)]{clune2019ai}
Jeff Clune.
\newblock Ai-gas: Ai-generating algorithms, an alternate paradigm for producing
  general artificial intelligence.
\newblock \emph{arXiv preprint arXiv:1905.10985}, 2019.

\bibitem[Colas et~al.(2020)Colas, Madhavan, Huizinga, and
  Clune]{colas2020scaling}
C{\'e}dric Colas, Vashisht Madhavan, Joost Huizinga, and Jeff Clune.
\newblock Scaling map-elites to deep neuroevolution.
\newblock In \emph{Proceedings of the 2020 Genetic and Evolutionary Computation
  Conference}, pages 67--75, 2020.

\bibitem[Collobert et~al.(2011)Collobert, Kavukcuoglu, and
  Farabet]{collobert2011torch}
Ronan Collobert, Koray Kavukcuoglu, and Clément Farabet.
\newblock Torch7: A matlab-like environment for machine learning.
\newblock 2011.
\newblock URL \url{http://infoscience.epfl.ch/record/192376}.

\bibitem[Coumans and Bai(2016--2020)]{coumans2020}
Erwin Coumans and Yunfei Bai.
\newblock Pybullet, a python module for physics simulation for games, robotics
  and machine learning.
\newblock \url{http://pybullet.org}, 2016--2020.

\bibitem[Cully(2019)]{cully2019autonomous}
Antoine Cully.
\newblock Autonomous skill discovery with quality-diversity and unsupervised
  descriptors.
\newblock In \emph{Proceedings of the Genetic and Evolutionary Computation
  Conference}, pages 81--89, 2019.

\bibitem[Cully(2020)]{cully2020multi}
Antoine Cully.
\newblock Multi-emitter map-elites: Improving quality, diversity and
  convergence speed with heterogeneous sets of emitters.
\newblock \emph{arXiv preprint arXiv:2007.05352}, 2020.

\bibitem[Cully and Demiris(2017)]{cully2017quality}
Antoine Cully and Yiannis Demiris.
\newblock Quality and diversity optimization: A unifying modular framework.
\newblock \emph{IEEE Transactions on Evolutionary Computation}, 22\penalty0
  (2):\penalty0 245--259, 2017.

\bibitem[Cully and Mouret(2013)]{cully2013behavioral}
Antoine Cully and Jean-Baptiste Mouret.
\newblock Behavioral repertoire learning in robotics.
\newblock In \emph{Proceedings of the 15th annual conference on Genetic and
  evolutionary computation}, pages 175--182, 2013.

\bibitem[Cully et~al.(2015)Cully, Clune, Tarapore, and Mouret]{cully2015robots}
Antoine Cully, Jeff Clune, Danesh Tarapore, and Jean-Baptiste Mouret.
\newblock Robots that can adapt like animals.
\newblock \emph{Nature}, 521\penalty0 (7553):\penalty0 503--507, 2015.

\bibitem[Deng et~al.(2009)Deng, Dong, Socher, Li, Li, and
  Fei-Fei]{deng2009imagenet}
Jia Deng, Wei Dong, Richard Socher, Li-Jia Li, Kai Li, and Li~Fei-Fei.
\newblock Imagenet: A large-scale hierarchical image database.
\newblock In \emph{2009 IEEE conference on computer vision and pattern
  recognition}, pages 248--255. Ieee, 2009.

\bibitem[Devlin et~al.(2018)Devlin, Chang, Lee, and Toutanova]{devlin2018bert}
Jacob Devlin, Ming-Wei Chang, Kenton Lee, and Kristina Toutanova.
\newblock Bert: Pre-training of deep bidirectional transformers for language
  understanding.
\newblock \emph{arXiv preprint arXiv:1810.04805}, 2018.

\bibitem[Dhar(2020)]{dhar2020carbon}
Payal Dhar.
\newblock The carbon impact of artificial intelligence.
\newblock \emph{Nature Machine Intelligence}, 2\penalty0 (8):\penalty0
  423--425, 2020.

\bibitem[Dobbe and Whittaker(2019)]{dobbe2019climate}
R.~Dobbe and M.~Whittaker.
\newblock Ai and climate change: How they’re connected, and what we can do
  about it.
\newblock 2019.

\bibitem[Earle et~al.(2021)Earle, Snider, Fontaine, Nikolaidis, and
  Togelius]{earle2021illuminating}
Sam Earle, Justin Snider, Matthew~C Fontaine, Stefanos Nikolaidis, and Julian
  Togelius.
\newblock Illuminating diverse neural cellular automata for level generation.
\newblock \emph{arXiv preprint arXiv:2109.05489}, 2021.

\bibitem[Ecoffet et~al.(2021)Ecoffet, Huizinga, Lehman, Stanley, and
  Clune]{ecoffet2021first}
Adrien Ecoffet, Joost Huizinga, Joel Lehman, Kenneth~O Stanley, and Jeff Clune.
\newblock First return, then explore.
\newblock \emph{Nature}, 590\penalty0 (7847):\penalty0 580--586, 2021.

\bibitem[Eysenbach et~al.(2018)Eysenbach, Gupta, Ibarz, and
  Levine]{eysenbach2018diversity}
Benjamin Eysenbach, Abhishek Gupta, Julian Ibarz, and Sergey Levine.
\newblock Diversity is all you need: Learning skills without a reward function.
\newblock \emph{arXiv preprint arXiv:1802.06070}, 2018.

\bibitem[Fontaine and Nikolaidis(2021)]{fontaine2021differentiable}
Matthew Fontaine and Stefanos Nikolaidis.
\newblock Differentiable quality diversity.
\newblock \emph{Advances in Neural Information Processing Systems}, 34, 2021.

\bibitem[Fontaine et~al.(2019)Fontaine, Lee, Soros, de~Mesentier~Silva,
  Togelius, and Hoover]{fontaine2019mapping}
Matthew~C Fontaine, Scott Lee, Lisa~B Soros, Fernando de~Mesentier~Silva,
  Julian Togelius, and Amy~K Hoover.
\newblock Mapping hearthstone deck spaces through map-elites with sliding
  boundaries.
\newblock In \emph{Proceedings of The Genetic and Evolutionary Computation
  Conference}, pages 161--169, 2019.

\bibitem[Fontaine et~al.(2020{\natexlab{a}})Fontaine, Liu, Khalifa, Modi,
  Togelius, Hoover, and Nikolaidis]{fontaine2020illuminating}
Matthew~C Fontaine, Ruilin Liu, Ahmed Khalifa, Jignesh Modi, Julian Togelius,
  Amy~K Hoover, and Stefanos Nikolaidis.
\newblock Illuminating mario scenes in the latent space of a generative
  adversarial network.
\newblock \emph{arXiv preprint arXiv:2007.05674}, 2020{\natexlab{a}}.

\bibitem[Fontaine et~al.(2020{\natexlab{b}})Fontaine, Togelius, Nikolaidis, and
  Hoover]{fontaine2020covariance}
Matthew~C Fontaine, Julian Togelius, Stefanos Nikolaidis, and Amy~K Hoover.
\newblock Covariance matrix adaptation for the rapid illumination of behavior
  space.
\newblock In \emph{Proceedings of the 2020 genetic and evolutionary computation
  conference}, pages 94--102, 2020{\natexlab{b}}.

\bibitem[Freeman et~al.(2021)Freeman, Frey, Raichuk, Girgin, Mordatch, and
  Bachem]{brax2021github}
C.~Daniel Freeman, Erik Frey, Anton Raichuk, Sertan Girgin, Igor Mordatch, and
  Olivier Bachem.
\newblock Brax - a differentiable physics engine for large scale rigid body
  simulation, 2021.
\newblock URL \url{http://github.com/google/brax}.

\bibitem[Gaier et~al.(2018)Gaier, Asteroth, and Mouret]{gaier2018data}
Adam Gaier, Alexander Asteroth, and Jean-Baptiste Mouret.
\newblock Data-efficient design exploration through surrogate-assisted
  illumination.
\newblock \emph{Evolutionary computation}, 26\penalty0 (3):\penalty0 381--410,
  2018.

\bibitem[Gravina et~al.(2019)Gravina, Khalifa, Liapis, Togelius, and
  Yannakakis]{gravina2019procedural}
Daniele Gravina, Ahmed Khalifa, Antonios Liapis, Julian Togelius, and
  Georgios~N Yannakakis.
\newblock Procedural content generation through quality diversity.
\newblock In \emph{2019 IEEE Conference on Games (CoG)}, pages 1--8. IEEE,
  2019.

\bibitem[Gu et~al.(2021)Gu, Diaz, Freeman, Furuta, Ghasemipour, Raichuk, David,
  Frey, Coumans, and Bachem]{gu2021braxlines}
Shixiang~Shane Gu, Manfred Diaz, Daniel~C Freeman, Hiroki Furuta, Seyed
  Kamyar~Seyed Ghasemipour, Anton Raichuk, Byron David, Erik Frey, Erwin
  Coumans, and Olivier Bachem.
\newblock Braxlines: Fast and interactive toolkit for rl-driven behavior
  engineering beyond reward maximization.
\newblock \emph{arXiv preprint arXiv:2110.04686}, 2021.

\bibitem[Hansen et~al.(2010)Hansen, Auger, Ros, Finck, and
  Posik]{hansen:hal-00545727}
Nikolaus Hansen, Anne Auger, Raymond Ros, Steffen Finck, and Petr Posik.
\newblock {Comparing Results of 31 Algorithms from the Black-Box Optimization
  Benchmarking BBOB-2009}.
\newblock In \emph{{ACM-GECCO Genetic and Evolutionary Computation
  Conference}}, Portland, United States, July 2010.
\newblock URL \url{https://hal.archives-ouvertes.fr/hal-00545727}.
\newblock pp. 1689-1696.

\bibitem[Hansen et~al.(2021)Hansen, Auger, Ros, Mersmann, Tu{\v{s}}ar, and
  Brockhoff]{hansen2021coco}
Nikolaus Hansen, Anne Auger, Raymond Ros, Olaf Mersmann, Tea Tu{\v{s}}ar, and
  Dimo Brockhoff.
\newblock Coco: A platform for comparing continuous optimizers in a black-box
  setting.
\newblock \emph{Optimization Methods and Software}, 36\penalty0 (1):\penalty0
  114--144, 2021.

\bibitem[He et~al.(2016)He, Zhang, Ren, and Sun]{he2016deep}
Kaiming He, Xiangyu Zhang, Shaoqing Ren, and Jian Sun.
\newblock Deep residual learning for image recognition.
\newblock In \emph{Proceedings of the IEEE conference on computer vision and
  pattern recognition}, pages 770--778, 2016.

\bibitem[Hochreiter and Schmidhuber(1997)]{hochreiter1997long}
Sepp Hochreiter and J{\"u}rgen Schmidhuber.
\newblock Long short-term memory.
\newblock \emph{Neural computation}, 9\penalty0 (8):\penalty0 1735--1780, 1997.

\bibitem[Jumper et~al.(2021)Jumper, Evans, Pritzel, Green, Figurnov,
  Ronneberger, Tunyasuvunakool, Bates, {\v{Z}}{\'\i}dek, Potapenko, Bridgland,
  Meyer, Kohl, Ballard, Cowie, Romera-Paredes, Nikolov, Jain, Adler, Back,
  Petersen, Reiman, Clancy, Zielinski, Steinegger, Pacholska, Berghammer,
  Bodenstein, Silver, Vinyals, Senior, Kavukcuoglu, Kohli, and
  Hassabis]{AlphaFold2021}
John Jumper, Richard Evans, Alexander Pritzel, Tim Green, Michael Figurnov,
  Olaf Ronneberger, Kathryn Tunyasuvunakool, Russ Bates, Augustin
  {\v{Z}}{\'\i}dek, Anna Potapenko, Alex Bridgland, Clemens Meyer, Simon A~A
  Kohl, Andrew~J Ballard, Andrew Cowie, Bernardino Romera-Paredes, Stanislav
  Nikolov, Rishub Jain, Jonas Adler, Trevor Back, Stig Petersen, David Reiman,
  Ellen Clancy, Michal Zielinski, Martin Steinegger, Michalina Pacholska, Tamas
  Berghammer, Sebastian Bodenstein, David Silver, Oriol Vinyals, Andrew~W
  Senior, Koray Kavukcuoglu, Pushmeet Kohli, and Demis Hassabis.
\newblock Highly accurate protein structure prediction with {AlphaFold}.
\newblock \emph{Nature}, 596\penalty0 (7873):\penalty0 583--589, 2021.
\newblock \doi{10.1038/s41586-021-03819-2}.

\bibitem[Kaushik et~al.(2020)Kaushik, Desreumaux, and
  Mouret]{kaushik2020adaptive}
Rituraj Kaushik, Pierre Desreumaux, and Jean-Baptiste Mouret.
\newblock Adaptive prior selection for repertoire-based online adaptation in
  robotics.
\newblock \emph{Frontiers in Robotics and AI}, 6:\penalty0 151, 2020.

\bibitem[Keller et~al.(2020)Keller, Tanneberg, Stark, and
  Peters]{keller2020model}
Leon Keller, Daniel Tanneberg, Svenja Stark, and Jan Peters.
\newblock Model-based quality-diversity search for efficient robot learning.
\newblock \emph{arXiv preprint arXiv:2008.04589}, 2020.

\bibitem[Krizhevsky et~al.(2012)Krizhevsky, Sutskever, and
  Hinton]{krizhevsky2012imagenet}
Alex Krizhevsky, Ilya Sutskever, and Geoffrey~E Hinton.
\newblock Imagenet classification with deep convolutional neural networks.
\newblock \emph{Advances in neural information processing systems},
  25:\penalty0 1097--1105, 2012.

\bibitem[Lacoste et~al.(2019)Lacoste, Luccioni, Schmidt, and
  Dandres]{lacoste2019quantifying}
Alexandre Lacoste, Alexandra Luccioni, Victor Schmidt, and Thomas Dandres.
\newblock Quantifying the carbon emissions of machine learning.
\newblock \emph{arXiv preprint arXiv:1910.09700}, 2019.

\bibitem[Lecun et~al.(2015)Lecun, Bengio, and Hinton]{lecun2015DL}
Yann Lecun, Yoshua Bengio, and Geoffrey Hinton.
\newblock Deep learning.
\newblock \emph{Nature 2015 521:7553}, 521:\penalty0 436--444, 5 2015.
\newblock ISSN 1476-4687.
\newblock \doi{10.1038/nature14539}.
\newblock URL \url{https://www.nature.com/articles/nature14539}.

\bibitem[Lee et~al.(2018)Lee, Grey, Ha, Kunz, Jain, Ye, Srinivasa, Stilman, and
  Liu]{Lee2018}
Jeongseok Lee, Michael~X. Grey, Sehoon Ha, Tobias Kunz, Sumit Jain, Yuting Ye,
  Siddhartha~S. Srinivasa, Mike Stilman, and C.~Karen Liu.
\newblock {DART}: Dynamic animation and robotics toolkit.
\newblock \emph{The Journal of Open Source Software}, 3\penalty0 (22):\penalty0
  500, Feb 2018.
\newblock \doi{10.21105/joss.00500}.
\newblock URL \url{https://doi.org/10.21105/joss.00500}.

\bibitem[Lehman and Stanley(2011{\natexlab{a}})]{lehman2011abandoning}
Joel Lehman and Kenneth~O Stanley.
\newblock Abandoning objectives: Evolution through the search for novelty
  alone.
\newblock \emph{Evolutionary computation}, 19\penalty0 (2):\penalty0 189--223,
  2011{\natexlab{a}}.

\bibitem[Lehman and Stanley(2011{\natexlab{b}})]{lehman2011evolving}
Joel Lehman and Kenneth~O Stanley.
\newblock Evolving a diversity of virtual creatures through novelty search and
  local competition.
\newblock In \emph{Proceedings of the 13th annual conference on Genetic and
  evolutionary computation}, pages 211--218, 2011{\natexlab{b}}.

\bibitem[Lim et~al.(2021)Lim, Grillotti, Bernasconi, and
  Cully]{lim2021dynamics}
Bryan Lim, Luca Grillotti, Lorenzo Bernasconi, and Antoine Cully.
\newblock Dynamics-aware quality-diversity for efficient learning of skill
  repertoires.
\newblock \emph{arXiv preprint arXiv:2109.08522}, 2021.

\bibitem[Makoviychuk et~al.(2021)Makoviychuk, Wawrzyniak, Guo, Lu, Storey,
  Macklin, Hoeller, Rudin, Allshire, Handa, and State]{makoviychuk2021isaac}
Viktor Makoviychuk, Lukasz Wawrzyniak, Yunrong Guo, Michelle Lu, Kier Storey,
  Miles Macklin, David Hoeller, Nikita Rudin, Arthur Allshire, Ankur Handa, and
  Gavriel State.
\newblock Isaac gym: High performance gpu-based physics simulation for robot
  learning.
\newblock \emph{CoRR}, abs/2108.10470, 2021.
\newblock URL \url{https://arxiv.org/abs/2108.10470}.

\bibitem[Mouret and Doncieux(2010)]{Mouret2010}
J.-B. Mouret and S.~Doncieux.
\newblock {SFERES}v2: Evolvin' in the multi-core world.
\newblock In \emph{Proc. of Congress on Evolutionary Computation (CEC)}, pages
  4079--4086, 2010.

\bibitem[Mouret and Doncieux(2012)]{mouret2012encouraging}
J-B Mouret and St{\'e}phane Doncieux.
\newblock Encouraging behavioral diversity in evolutionary robotics: An
  empirical study.
\newblock \emph{Evolutionary computation}, 20\penalty0 (1):\penalty0 91--133,
  2012.

\bibitem[Mouret and Clune(2015)]{mouret2015illuminating}
Jean-Baptiste Mouret and Jeff Clune.
\newblock Illuminating search spaces by mapping elites.
\newblock \emph{arXiv preprint arXiv:1504.04909}, 2015.

\bibitem[Mouret and Doncieux(2009)]{mouret2009overcoming}
Jean-Baptiste Mouret and St{\'e}phane Doncieux.
\newblock Overcoming the bootstrap problem in evolutionary robotics using
  behavioral diversity.
\newblock In \emph{2009 IEEE Congress on Evolutionary Computation}, pages
  1161--1168. IEEE, 2009.

\bibitem[Nilsson and Cully(2021)]{nilsson2021policy}
Olle Nilsson and Antoine Cully.
\newblock Policy gradient assisted map-elites.
\newblock In \emph{Proceedings of the Genetic and Evolutionary Computation
  Conference}, pages 866--875, 2021.

\bibitem[Paolo et~al.(2020)Paolo, Laflaquiere, Coninx, and
  Doncieux]{paolo2020unsupervised}
Giuseppe Paolo, Alban Laflaquiere, Alexandre Coninx, and Stephane Doncieux.
\newblock Unsupervised learning and exploration of reachable outcome space.
\newblock In \emph{2020 IEEE International Conference on Robotics and
  Automation (ICRA)}, pages 2379--2385. IEEE, 2020.

\bibitem[Paszke et~al.(2019)Paszke, Gross, Massa, Lerer, Bradbury, Chanan,
  Killeen, Lin, Gimelshein, Antiga, Desmaison, K{\"{o}}pf, Yang, DeVito,
  Raison, Tejani, Chilamkurthy, Steiner, Fang, Bai, and
  Chintala]{paszke2019pytorch}
Adam Paszke, Sam Gross, Francisco Massa, Adam Lerer, James Bradbury, Gregory
  Chanan, Trevor Killeen, Zeming Lin, Natalia Gimelshein, Luca Antiga, Alban
  Desmaison, Andreas K{\"{o}}pf, Edward~Z. Yang, Zach DeVito, Martin Raison,
  Alykhan Tejani, Sasank Chilamkurthy, Benoit Steiner, Lu~Fang, Junjie Bai, and
  Soumith Chintala.
\newblock Pytorch: An imperative style, high-performance deep learning library.
\newblock \emph{CoRR}, abs/1912.01703, 2019.
\newblock URL \url{http://arxiv.org/abs/1912.01703}.

\bibitem[Pierrot et~al.(2021)Pierrot, Mac{\'e}, Cideron, Beguir, Cully, Sigaud,
  and Perrin]{pierrot2021diversity}
Thomas Pierrot, Valentin Mac{\'e}, Geoffrey Cideron, Karim Beguir, Antoine
  Cully, Olivier Sigaud, and Nicolas Perrin.
\newblock Diversity policy gradient for sample efficient quality-diversity
  optimization.
\newblock 2021.

\bibitem[Pugh et~al.(2016)Pugh, Soros, and Stanley]{pugh2016quality}
Justin~K Pugh, Lisa~B Soros, and Kenneth~O Stanley.
\newblock Quality diversity: A new frontier for evolutionary computation.
\newblock \emph{Frontiers in Robotics and AI}, 3:\penalty0 40, 2016.

\bibitem[Raina et~al.(2009)Raina, Madhavan, and Ng]{raina2009gpu}
Rajat Raina, Anand Madhavan, and Andrew~Y. Ng.
\newblock Large-scale deep unsupervised learning using graphics processors.
\newblock In \emph{Proceedings of the 26th Annual International Conference on
  Machine Learning}, ICML '09, page 873–880, New York, NY, USA, 2009.
  Association for Computing Machinery.
\newblock ISBN 9781605585161.
\newblock \doi{10.1145/1553374.1553486}.
\newblock URL \url{https://doi.org/10.1145/1553374.1553486}.

\bibitem[Redmon et~al.(2016)Redmon, Divvala, Girshick, and
  Farhadi]{redmon2016you}
Joseph Redmon, Santosh Divvala, Ross Girshick, and Ali Farhadi.
\newblock You only look once: Unified, real-time object detection.
\newblock In \emph{Proceedings of the IEEE conference on computer vision and
  pattern recognition}, pages 779--788, 2016.

\bibitem[Rudin et~al.(2021)Rudin, Hoeller, Reist, and
  Hutter]{rudin2021learning}
Nikita Rudin, David Hoeller, Philipp Reist, and Marco Hutter.
\newblock Learning to walk in minutes using massively parallel deep
  reinforcement learning.
\newblock \emph{arXiv preprint arXiv:2109.11978}, 2021.

\bibitem[Schulman et~al.(2017)Schulman, Wolski, Dhariwal, Radford, and
  Klimov]{schulman2017proximal}
John Schulman, Filip Wolski, Prafulla Dhariwal, Alec Radford, and Oleg Klimov.
\newblock Proximal policy optimization algorithms.
\newblock \emph{arXiv preprint arXiv:1707.06347}, 2017.

\bibitem[Schwartz et~al.(2020)Schwartz, Dodge, Smith, and
  Etzioni]{schwartz2020green}
Roy Schwartz, Jesse Dodge, Noah~A Smith, and Oren Etzioni.
\newblock Green ai.
\newblock \emph{Communications of the ACM}, 63\penalty0 (12):\penalty0 54--63,
  2020.

\bibitem[Shoeybi et~al.(2019)Shoeybi, Patwary, Puri, LeGresley, Casper, and
  Catanzaro]{shoeybi2019megatron}
Mohammad Shoeybi, Mostofa Patwary, Raul Puri, Patrick LeGresley, Jared Casper,
  and Bryan Catanzaro.
\newblock Megatron-lm: Training multi-billion parameter language models using
  model parallelism.
\newblock \emph{arXiv preprint arXiv:1909.08053}, 2019.

\bibitem[Smith et~al.(2022)Smith, Patwary, Norick, LeGresley, Rajbhandari,
  Casper, Liu, Prabhumoye, Zerveas, Korthikanti, Zhang, Child, Aminabadi,
  Bernauer, Song, Shoeybi, He, Houston, Tiwary, and Catanzaro]{smith2022using}
Shaden Smith, Mostofa Patwary, Brandon Norick, Patrick LeGresley, Samyam
  Rajbhandari, Jared Casper, Zhun Liu, Shrimai Prabhumoye, George Zerveas,
  Vijay Korthikanti, Elton Zhang, Rewon Child, Reza~Yazdani Aminabadi, Julie
  Bernauer, Xia Song, Mohammad Shoeybi, Yuxiong He, Michael Houston, Saurabh
  Tiwary, and Bryan Catanzaro.
\newblock Using deepspeed and megatron to train megatron-turing nlg 530b, a
  large-scale generative language model, 2022.

\bibitem[Stanley(2019)]{stanley2019open}
Kenneth~O Stanley.
\newblock Why open-endedness matters.
\newblock \emph{Artificial life}, 25\penalty0 (3):\penalty0 232--235, 2019.

\bibitem[Stanley et~al.(2017)Stanley, Lehman, and Soros]{stanley2017open}
Kenneth~O Stanley, Joel Lehman, and Lisa Soros.
\newblock Open-endedness: The last grand challenge you’ve never heard of.
\newblock \emph{While open-endedness could be a force for discovering
  intelligence, it could also be a component of AI itself}, 2017.

\bibitem[Steinkrau et~al.(2005)Steinkrau, Simard, and Buck]{steinkrau205gpu}
Dave Steinkrau, Patrice~Y. Simard, and Ian Buck.
\newblock Using gpus for machine learning algorithms.
\newblock In \emph{Proceedings of the Eighth International Conference on
  Document Analysis and Recognition}, ICDAR '05, page 1115–1119, USA, 2005.
  IEEE Computer Society.
\newblock ISBN 0769524206.
\newblock \doi{10.1109/ICDAR.2005.251}.
\newblock URL \url{https://doi.org/10.1109/ICDAR.2005.251}.

\bibitem[Tjanaka et~al.(2021)Tjanaka, Fontaine, Zhang, Sommerer, Dennler, and
  Nikolaidis]{pyribs}
Bryon Tjanaka, Matthew~C. Fontaine, Yulun Zhang, Sam Sommerer, Nathan Dennler,
  and Stefanos Nikolaidis.
\newblock pyribs: A bare-bones python library for quality diversity
  optimization.
\newblock \url{https://github.com/icaros-usc/pyribs}, 2021.

\bibitem[Tjanaka et~al.(2022)Tjanaka, Fontaine, Togelius, and
  Nikolaidis]{tjanaka2022approximating}
Bryon Tjanaka, Matthew~C Fontaine, Julian Togelius, and Stefanos Nikolaidis.
\newblock Approximating gradients for differentiable quality diversity in
  reinforcement learning.
\newblock \emph{arXiv preprint arXiv:2202.03666}, 2022.

\bibitem[Todorov et~al.(2012)Todorov, Erez, and Tassa]{todorov2012mujoco}
Emanuel Todorov, Tom Erez, and Yuval Tassa.
\newblock Mujoco: A physics engine for model-based control.
\newblock In \emph{2012 IEEE/RSJ International Conference on Intelligent Robots
  and Systems}, pages 5026--5033. IEEE, 2012.

\bibitem[Vassiliades and Mouret(2018)]{vassiliades2018iso}
Vassilis Vassiliades and Jean-Baptiste Mouret.
\newblock {Discovering the Elite Hypervolume by Leveraging Interspecies
  Correlation}.
\newblock In \emph{{GECCO 2018 - Genetic and Evolutionary Computation
  Conference}}, Kyoto, Japan, July 2018.
\newblock \doi{10.1145/3205455.3205602}.
\newblock URL \url{https://hal.inria.fr/hal-01764739}.

\bibitem[Vassiliades et~al.(2017)Vassiliades, Chatzilygeroudis, and
  Mouret]{vassiliades2017using}
Vassilis Vassiliades, Konstantinos Chatzilygeroudis, and Jean-Baptiste Mouret.
\newblock Using centroidal voronoi tessellations to scale up the
  multidimensional archive of phenotypic elites algorithm.
\newblock \emph{IEEE Transactions on Evolutionary Computation}, 22\penalty0
  (4):\penalty0 623--630, 2017.

\bibitem[Vaswani et~al.(2017)Vaswani, Shazeer, Parmar, Uszkoreit, Jones, Gomez,
  Kaiser, and Polosukhin]{vaswani2017attention}
Ashish Vaswani, Noam Shazeer, Niki Parmar, Jakob Uszkoreit, Llion Jones,
  Aidan~N Gomez, {\L}ukasz Kaiser, and Illia Polosukhin.
\newblock Attention is all you need.
\newblock In \emph{Advances in neural information processing systems}, pages
  5998--6008, 2017.

\bibitem[Wang et~al.(2019)Wang, Lehman, Clune, and Stanley]{wang2019poet}
Rui Wang, Joel Lehman, Jeff Clune, and Kenneth~O Stanley.
\newblock Poet: open-ended coevolution of environments and their optimized
  solutions.
\newblock In \emph{Proceedings of the Genetic and Evolutionary Computation
  Conference}, pages 142--151, 2019.

\bibitem[Wang et~al.(2020)Wang, Lehman, Rawal, Zhi, Li, Clune, and
  Stanley]{wang2020enhanced}
Rui Wang, Joel Lehman, Aditya Rawal, Jiale Zhi, Yulun Li, Jeffrey Clune, and
  Kenneth Stanley.
\newblock Enhanced poet: Open-ended reinforcement learning through unbounded
  invention of learning challenges and their solutions.
\newblock In \emph{International Conference on Machine Learning}, pages
  9940--9951. PMLR, 2020.

\end{thebibliography}
}



\newpage
\appendix

\section{Implementation Details}
In order to leverage the benefits of vectorised operations, JAX requires all the data on the device to be in the form of a Tensor with immutable size (classes and objects cannot be used directly).

\subsection{Archive Structure.}
The first data structure that we have to place on the device is the archive of MAP-Elites. However the number of solutions stored in the archive is meant to change overtime as new solutions are added to it and other solutions being replaced.
To accommodate these requirements, we implemented MAP-Elites with a fixed grid-sized archive in the form of a preallocated static array containing solutions with NaN values.
In MAP-Elites, the algorithm needs to store three attributes per solution as mentioned in section \ref{subsec:methods-ME}: (i) the objective values, (ii) the descriptors and (iii) the parameters. The objective values for all the solutions are initialized as an array of NaN values and the parameters for all the solutions as an array of zeros. The size of these arrays are defined as the number of cells the descriptor space contains.
The descriptors of the solutions are then directly encoded as indexes of the array storing the corresponding parameters and objective values. 
Defining the objective values and parameters as static arrays allows JAX to execute the addition and replacements of the new solutions directly on the device. This prevents the overhead caused by the transfer of data between the device and the CPU at each epoch to perform these operations.

\subsection{Solution Generation: Selection and Variation.}
\label{sec:appendix:selection-mutation}
As the archive is already completely preallocated with zero-outed solutions, sampling from the archive requires some considerations. To find cells that are actually filled, the stored objective values are used to check for valid, non-NaN entries. The indexes of these valid entries are stored at the top of a new static array of the size of the total number of cells. The bottom of this array is padded with zero values. To select solutions from the archive, the algorithm samples uniformly with replacement the indexes of valid content of this array to generate the indexes of the selected solutions $\params$. Consequently, this allows the algorithm to use these indexes to form a static array containing the solution parameters $\params$ of the selected parents by copying the corresponding solution parameters $\params$ from the archive.
Finally, the "Iso+LineDD" variation operator (see Algo.~\ref{algo:iso-line}) is mapped to the static batch of parents to generate the offspring. This operation is performed in parallel, since every pair of solutions used for the cross-over are independent.

\setlength{\intextsep}{10pt}%

\begin{algorithm}[H]
  \small
  \caption{Iso-line Variation Operator \cite{vassiliades2018iso}}
  \label{algo:iso-line}
  
  \newcommand\algorithmicitemindent{\hspace*{\algorithmicindent}\hspace*{\algorithmicindent}}
  \newcommand{\normal}{\mathcal{N}}
  
  \algnewcommand\algorithmicforeach{\textbf{for each}}
\algdef{S}[FOR]{ForEach}[1]{\algorithmicforeach\ #1\ \algorithmicdo}

  \begin{algorithmic}[1]
\Procedure{variation}{$(\params^{(1)}_j, \params^{(2)}_j)_{j\in\interval}$}
    \For{$j\in \interval$}
        \State{Sample: $\vec \epsilon_1 \sim \normal(\vec 0, \vec I),\quad \epsilon_2 \sim \normal(0, 1)$}
        \State{$\variaparams_j \leftarrow \params^{(1)}_j + \sigma_1 \vec \epsilon_1 + \sigma_2 \epsilon_{2} (\params^{(2)}_j - \params^{(1)}_j)$}
    \EndFor
    \Return{$(\variaparams_j)_{j\in\interval}$}
\EndProcedure
  \end{algorithmic}
\end{algorithm}

\subsection{Evaluation.}
The evaluation of the solutions is done in parallel as a batch with Brax ~\citep{brax2021github} as described in the previous section \ref{sec:hardware_acc_pop_methods}. 
Brax systematically simulates a static number of steps as it is built upon JAX and static arrays.
Due to the constraint of static arrays, Brax has to simulate all the steps in the specified episode length even if the agent fails during this episode. 
This differs from other simulators where the simulation stops as soon as an agent fails.
For this reason, we capture the descriptor from the trajectory up until the time step just before the agent fails the task. 
This is in contrast to taking the descriptor from the trajectory up until the final time step. 
Alternatively, we can also flag failed solutions with a "death flag" to discard them from being considered for addition.

\subsection{Archive Addition.}
Adding solutions to the archive is challenging since the batch of evaluated solutions can contain several solutions with the same descriptor but with different objective values. This causes a conflict when trying to apply the addition condition to all the solutions in a single JAX operation, as new solutions have to be compared to the content of the archive and between themselves. 
To overcome this issue, the archive addition follows four steps: (i) a boolean mask is applied to remove the new solutions with a death flag, (ii) the objective function of all new solutions having the same descriptor is set to $-\infty$ except for the best solution which keeps its objective value, (iii) we compare the objective value of existing solutions in the archive to the offspring and mask the new solutions having a worse objective, (iv) the boolean mask and the death flags  determine together which solution will be added by making JAX ignore solutions that haven't met the criteria to enter the archive.

All of these steps are JIT compatible since all the array sizes are static. As a result, the comparison and addition can be executed in a single set of operations (one for each of the four steps above) regardless of the actual batch size and directly on the device.

\section{Experimental Details}

\subsection{Metrics}
We also evaluate other commonly used metrics in QD-literature other than the QD-score.
The \textit{Coverage} is used as a measure of the diversity of solutions discovered in the archive without considering the performance of solutions. It is computed by counting the number of cells of the archive filled. 
The \textit{Best Objective} value metric captures the highest performing solution in the archive regardless of descriptor.
We present these metrics along with the QD-score for each environment in Appendix~\ref{section:suppl-results}.

\subsection{Tasks and Environments}

\newcommand{\sizeparam}{{N_{\Theta}}}

\textbf{Rastrigin and Spheres}
The full equations for the objective $f(\cdot)$ and descriptor functions $d(\cdot)$ for the black-box optimization tasks of rastrigin and spheres are given below, for all $\params\in\Theta\subseteq \R^{N_{\Theta}}$ (in all our experiments, we take: $\sizeparam=100$):
\begin{align}
    f_{rastrigin} (\params) &= -10 \sizeparam - \sum_{i=1}^\sizeparam  \left( \theta_i^2 - 10\cos(2\pi\theta_i) \right) \\
    f_{sphere} (\params) &= - \| \params \|^2_2
\end{align}
The descriptors in both optimization tasks are two-dimensional, and are represented by the first two parameters in the parameter vector $\params$. 
\begin{equation}
    d (\params) = 
    \begin{pmatrix}
        \theta_{0} \\
        \theta_{1}
    \end{pmatrix}
\end{equation}

\textbf{Continuous Control QD-RL}
The state and action space and the corresponding number of parameters of the policies used in the QD-RL environments are given in Table \ref{tab:tasks}. The QD related parameters such as the descriptor dimensions and number of cells are also provided.

As explained in Section \ref{sec:experiments}, the goal in omni-directional tasks is to discover policies to walk in every direction in an efficient manner. The objective $f_{omni}$ and descriptor $d_{omni}$ functions are given as:
\begin{align}
    f_{omni} &= \sum_{t=1}^{T}{r_{survive} + (-r_{torque})} \\
    d_{omni} &= 
    \begin{pmatrix}
        x_{T} \\
        y_{T}
    \end{pmatrix}
\end{align}
In the uni-directional tasks, the goal is to discover policies with diverse gaits that all walk as fast as possible~\citep{cully2015robots, nilsson2021policy}. The objective $f_{uni}$ and descriptor $d_{uni}$ functions are given as: 
\begin{align}
    f_{uni} &= \sum_{t=1}^{T}{r_{forward} + r_{survive} + (-r_{torque})} \\
    d_{uni} &= \frac{1}{T} \sum_{t=1}^{T}{
    \begin{pmatrix}
    C_1(t) \\
    \vdots \\
    C_I(t) 
    \end{pmatrix}\textrm{, with $I$ the number of feet.}
    }
\end{align}

\begin{table*}
\centering
  \begin{tabular}{ l | ccc | c }
    \toprule
    \multirow{2}{*}{QD-RL Task} & \multicolumn{3}{c}{Uni-directional} & \multicolumn{1}{c}{Omni-directional} \\
    & Walker& Ant & Humanoid & Ant \\
    \midrule
    \addlinespace
    State Space & 20 & 87 & 299 & 87 \\
    Action Space & 6 & 8 & 17 & 8 \\
    BD Dimensions & 2 & 4 & 2 & 2\\
    Number of cells & $40^2$ & $5^4$ & $40^2$ & $100^2$ \\
    \bottomrule
\end{tabular}
  \caption{Summary of uni-directional and omni-directional tasks across three environments (Walker, Ant, Humanoid) to benchmark QD algorithms applied to neuroevolution for RL domains.
  In the last line, the notation $x^y$ gives the number of cells in the corresponding MAP-Elites archive, where $y$ corresponds to the dimensionality of the descriptor and $x$ to the number of discretizations of each descriptor dimension.}
  \label{tab:tasks}
\end{table*}

\subsection{Computational Resources}
For all the batch size ablations on QDax (Section \ref{subsec:eff_of_massive_par}), we use a Google Cloud virtual machine running a single A100 GPU.

Experiments were conducted using Google Cloud Platform in region europe-west4, which has a carbon efficiency of 0.57 kgCO$_2$eq/kWh. A cumulative of 264 hours of computation was performed on hardware of type A100 PCIe 40/80GB (TDP of 250W). 
Total emissions are estimated to be 37.62 kgCO$_2$eq of which 100 percents were directly offset by the cloud provider.
Estimations were conducted using the \href{https://mlco2.github.io/impact#compute}{MachineLearning Impact calculator} presented in \cite{lacoste2019quantifying}.

\section{Supplementary Results}
\label{section:suppl-results}
\begin{figure}[h]
  \centering
  \includegraphics[width=\linewidth]{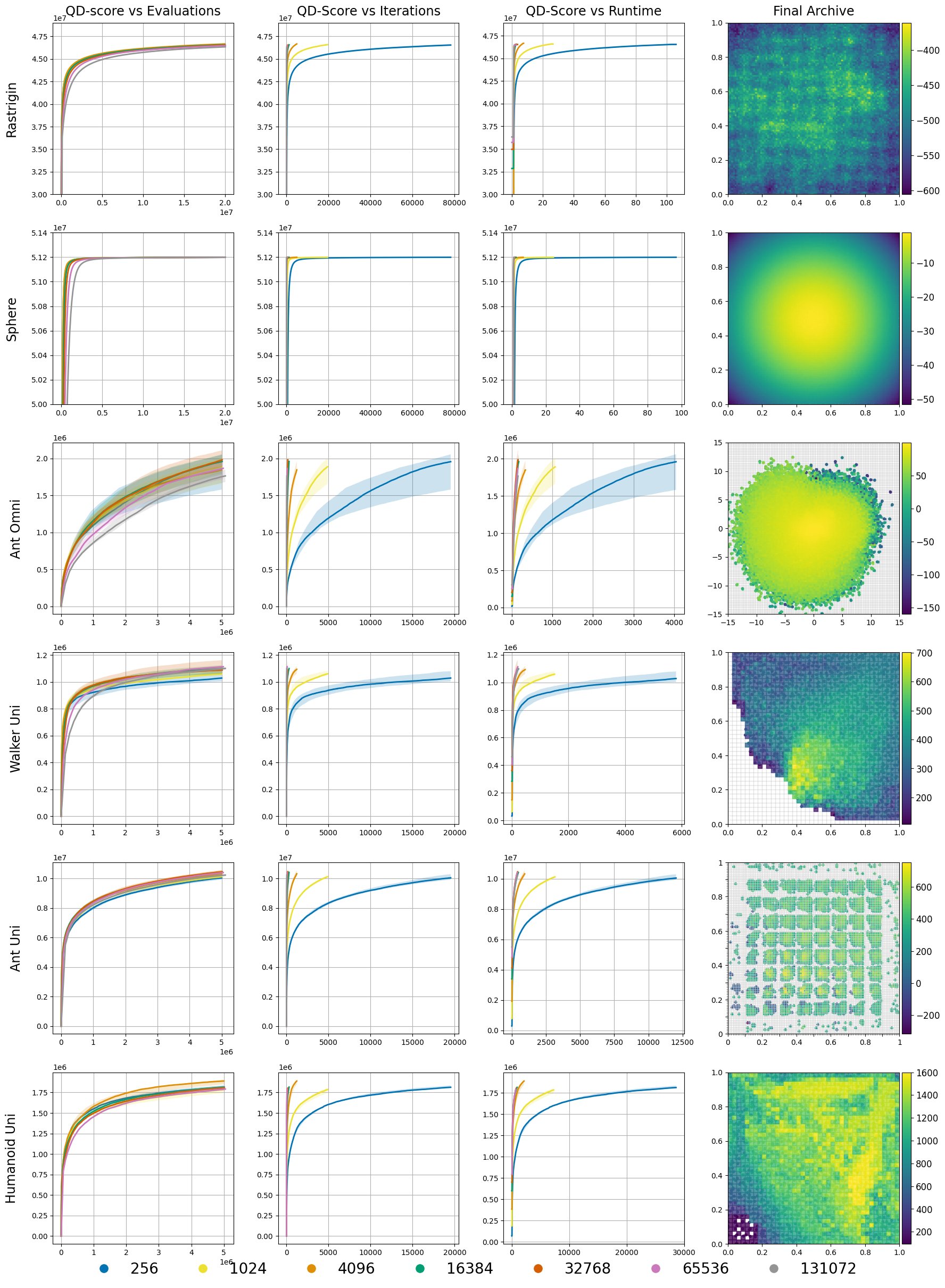}
  \caption{Performance curves for the QD-score of QDax for different batch sizes across generations of all environments. Compared to Fig. \ref{fig:all_envs_results} which removed the smaller batch sizes to increase readability of the larger batch sizes for the uni-directional tasks, we show all the batch sizes including 256 and 1024 for all the environments in this figure.
  We also show the Sphere optimization task here which was removed from Fig. \ref{fig:all_envs_results} due to space constraints.
  }
  \label{fig:results_all_envs_full}
\end{figure}

This section provides detailed results for all extended performance metrics (QD-Score, Coverage, Best Fitness) plot against evaluations, iterations and runtime. We provide this for each environment-task pair considered. 
Figure \ref{fig:results_all_envs_full} shows the performance of the difference batch sizes across all the environments.

\subsection{QD-RL}
Figures 6-9 show the extended results for each QD-RL environment-task pair.
\begin{figure}[h]
  \centering
  \includegraphics[width=\linewidth]{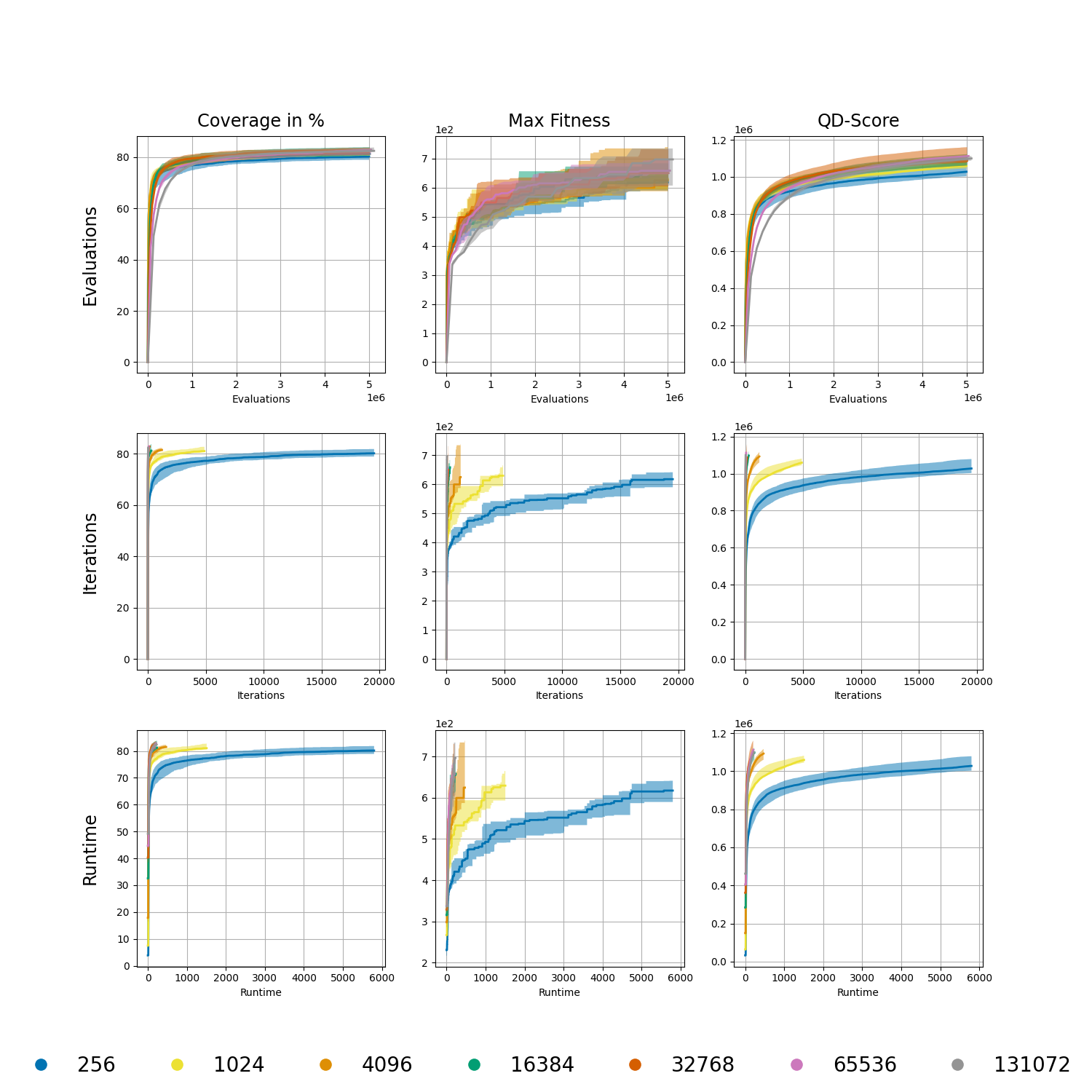}
  \caption{Performance curves of QDax for different batch sizes across generations of the algorithm on the \textbf{uni-directional Walker2d} task. The metrics are plotted against evaluations (top row), iterations (middle row) and runtime (bottom row) to demonstrate the effect of massive parallelization.}
  \label{fig:results_walker_uni}
\end{figure}

\begin{figure}[h]
  \centering
  \includegraphics[width=\linewidth]{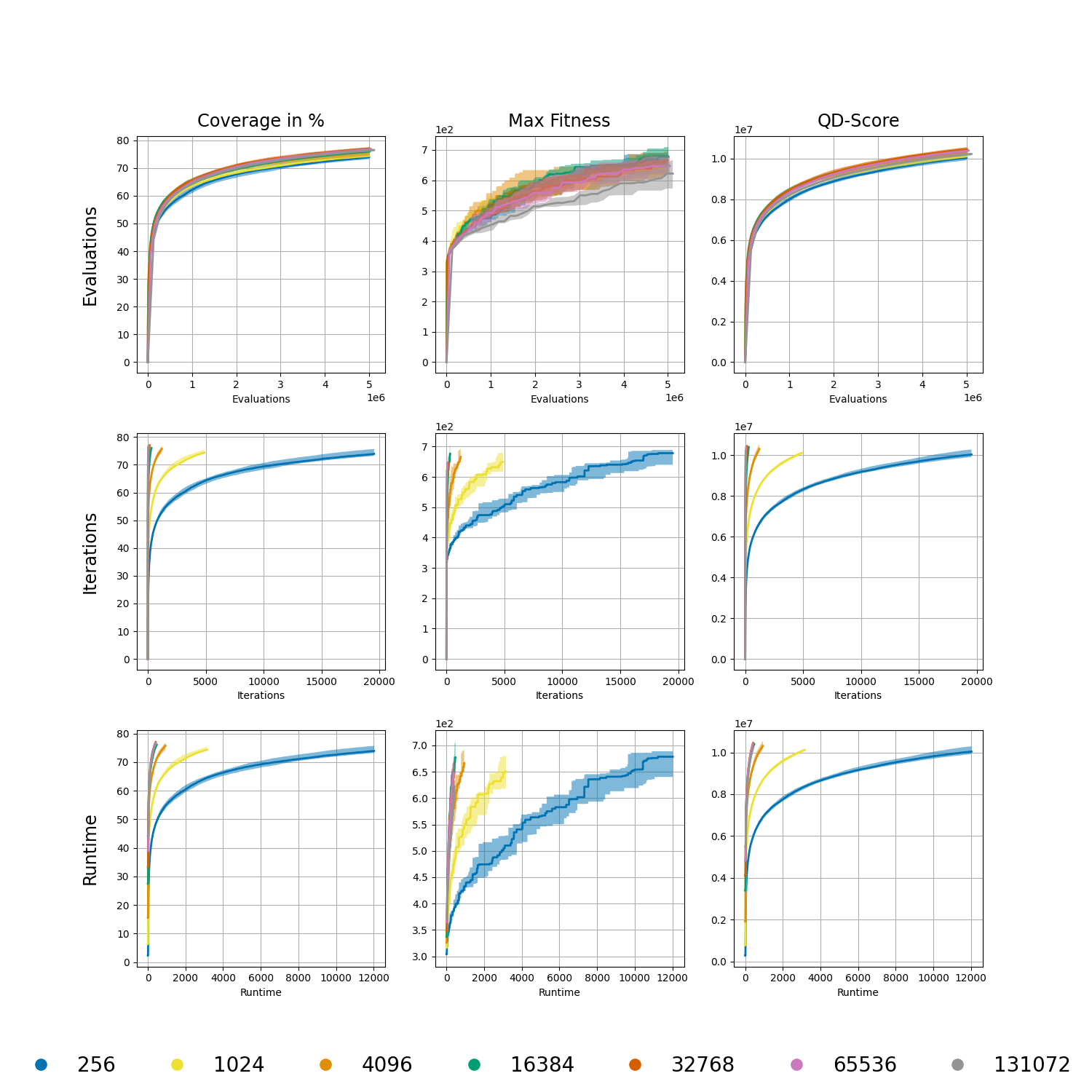}
  \caption{Performance curves of QDax for different batch sizes across generations of the algorithm on the \textbf{uni-directional Ant} task. The metrics are plotted against evaluations (top row), iterations (middle row) and runtime (bottom row) to demonstrate the effect of massive parallelization.}
  \label{fig:results_ant_uni}
\end{figure}

\begin{figure}[h]
  \centering
  \includegraphics[width=\linewidth]{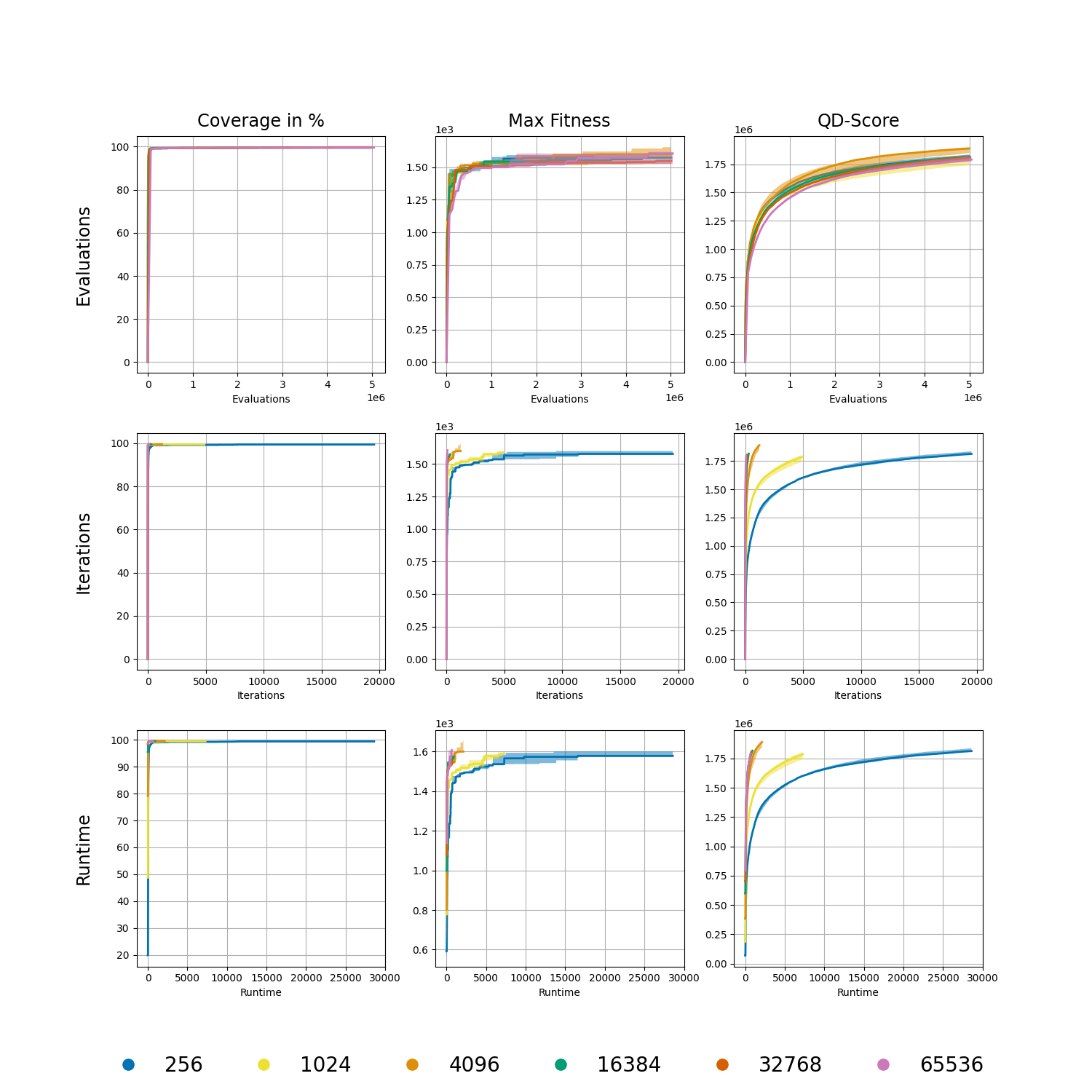}
  \caption{Performance curves of QDax for different batch sizes across generations of the algorithm on the \textbf{uni-directional Humanoid} task. The metrics are plotted against evaluations (top row), iterations (middle row) and runtime (bottom row) to demonstrate the effect of massive parallelization.}
  \label{fig:results_humanoid_uni}
\end{figure}

\begin{figure}[h]
  \centering
  \includegraphics[width=\linewidth]{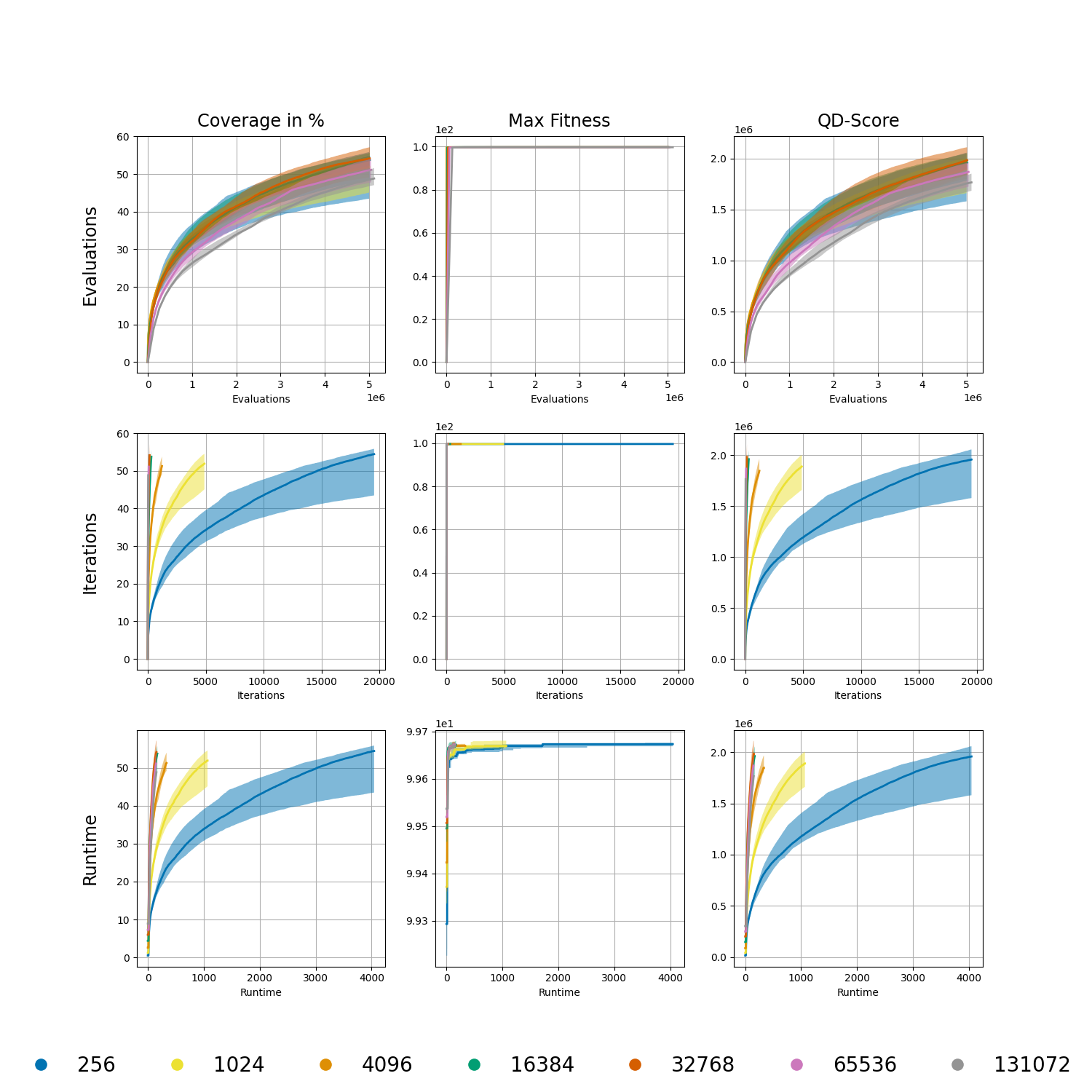}
  \caption{Performance curves of QDax for different batch sizes across generations of the algorithm on the \textbf{omni-directional Ant} task. The metrics are plotted against evaluations (top row), iterations (middle row) and runtime (bottom row) to demonstrate the effect of massive parallelization.}
  \label{fig:results_ant_omni}
\end{figure}

\subsection{Rastrigin and Sphere}
Figures 10-11 show the extended results for the rastrigin and sphere optimization task.
\begin{figure}[h]
  \centering
  \includegraphics[width=\linewidth]{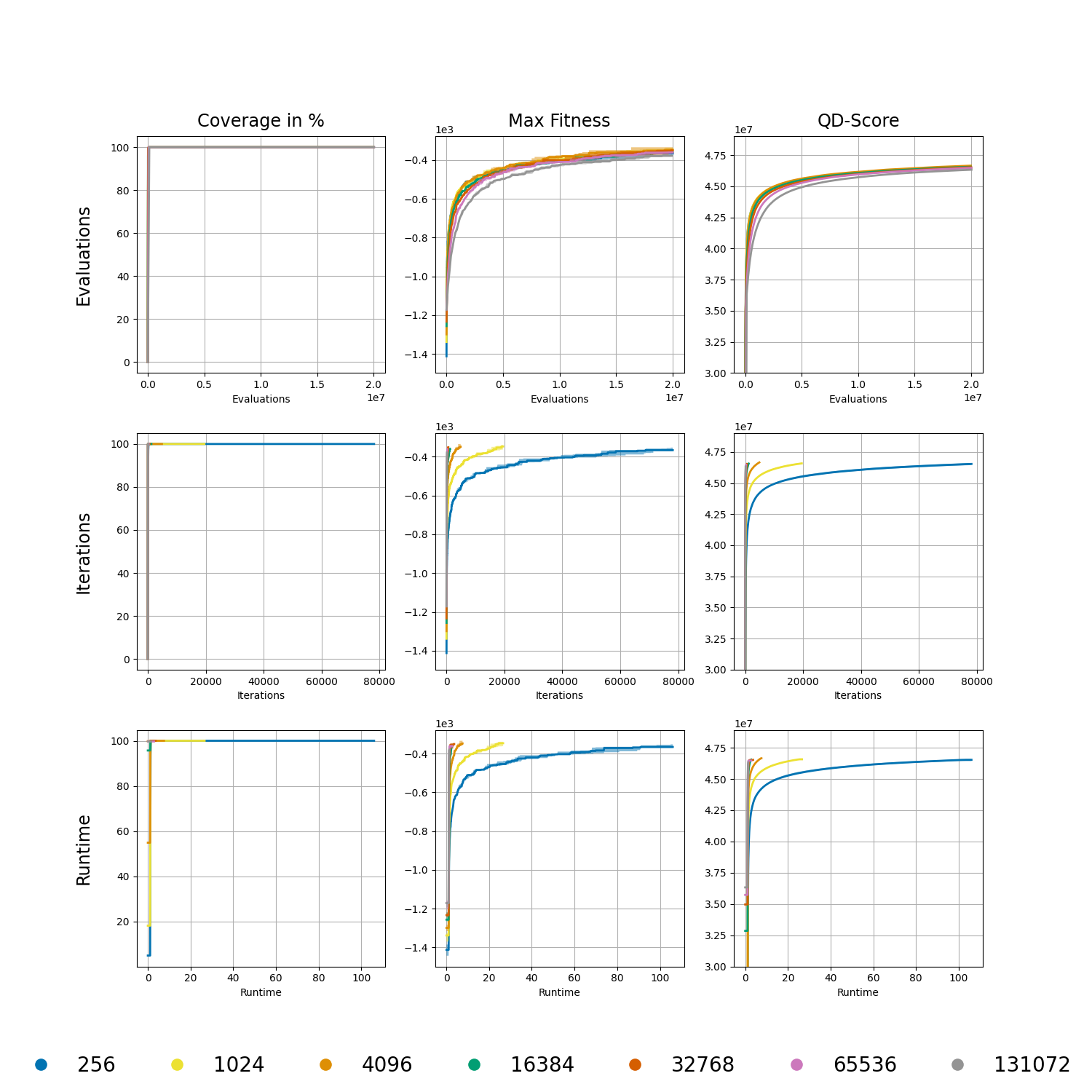}
  \caption{Performance curves of QDax for different batch sizes across generations of the algorithm on the \textbf{Rastrigin} optimization task. The metrics are plotted against evaluations (top row), iterations (middle row) and runtime (bottom row) to demonstrate the effect of massive parallelization.}
  \label{fig:results_rastrigin}
\end{figure}

\begin{figure}[h]
  \centering
  \includegraphics[width=\linewidth]{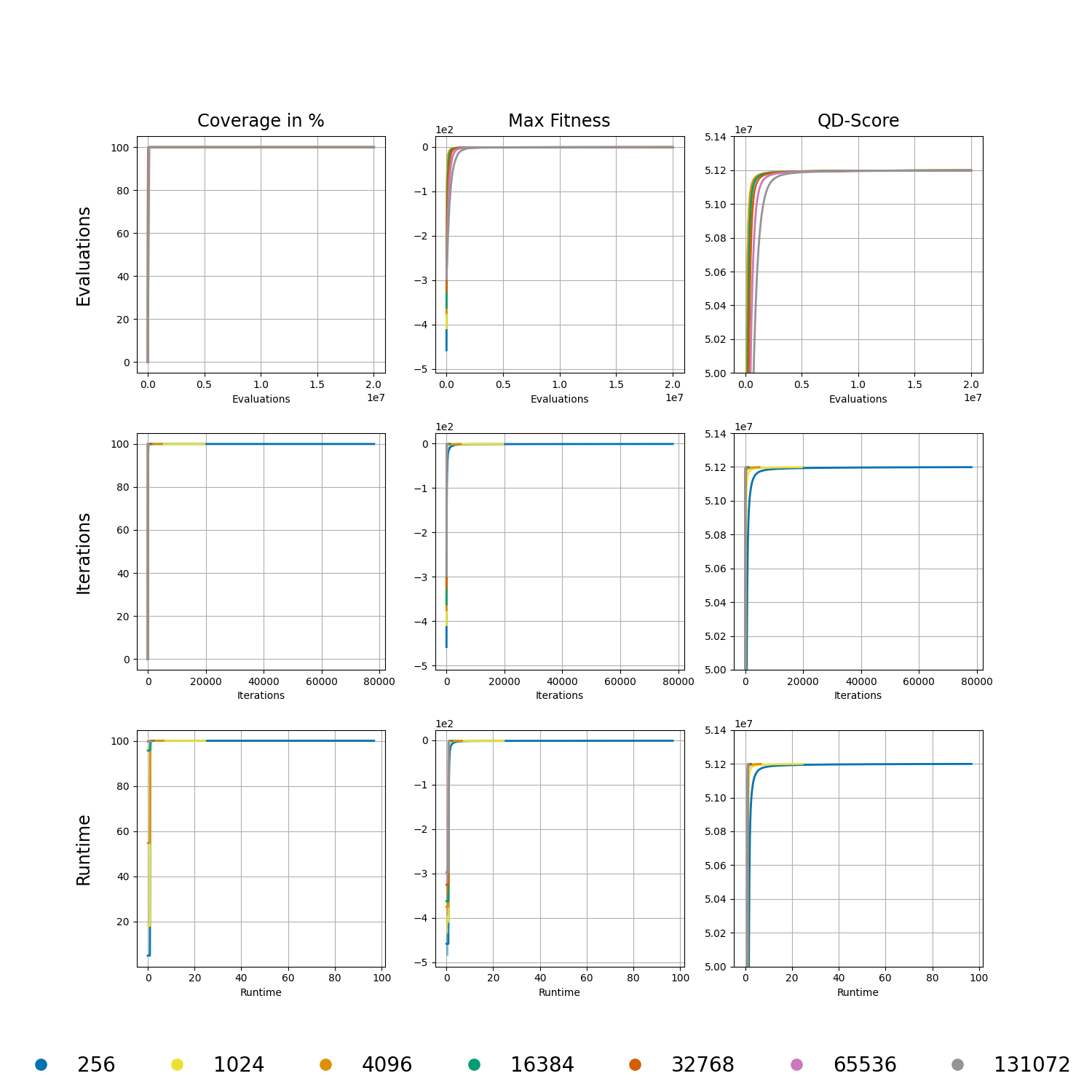}
  \caption{Performance curves of QDax for different batch sizes across generations of the algorithm on the \textbf{Sphere} optimization task. The metrics are plotted against evaluations (top row), iterations (middle row) and runtime (bottom row) to demonstrate the effect of massive parallelization.}
  \label{fig:results_sphere}
\end{figure}


\end{document}